\documentclass[a4paper]{cas-dc}

\usepackage[authoryear]{natbib}

\def\tsc#1{\csdef{#1}{\textsc{\lowercase{#1}}\xspace}}
\tsc{WGM}
\tsc{QE}
\usepackage{graphicx}
\usepackage{subcaption}

\begin{document}
\let\WriteBookmarks\relax
\def\floatpagepagefraction{1}
\def\textpagefraction{.001}
\let\printorcid\relax    

\shorttitle{BiMoRec}    

\shortauthors{Shi M,Chen L}  

\title [mode = title]{Medication Recommendation via Dual Molecular Modalities and Multi-Step Enhancement}

\author[1]{Shi Mu}
\ead{mushi@stumail.ysu.edu.cn}
\credit{Conceptualization, Methodology, Software, Validation,  Formal analysis, Data Curation, Investigation, Writing – original draft, Writing - Review \& Editing, Visualization}
\cormark[1]

\author[1]{Chen Li}
\ead{lichen36211@gmail.com}
\credit{Writing – original draft, Writing - Review \& Editing.}
\cormark[1]
\cortext[1]{First Author and Second Author contribute equally to this work.}

\author[1]{Xiang Li}
\ead{lilixiang222@gmail.com}
\credit{Writing - Review \& Editing}

\author[1,2]{Shunpan Liang}
\ead{liangshunpan@ysu.edu.cn}
\credit{Supervision, Funding acquisition}
\cormark[2]
\cortext[2]{Corresponding author.}


\affiliation[1]{
            organization={School of Information Science and Engineering, Yanshan University},
            city={Qinhuangdao},
            postcode={066004}, 
            country={China}
}

\affiliation[2]{
            organization={School of Information Science and Engineering, Xinjiang College of Science \& Technology},
            city={Korla},
            postcode={841000}, 
            country={China}
}





\begin{abstract}
As the integration of artificial intelligence technology with the medical field deepens, medication recommendation, as an important subfield, demonstrates immense potential value. Medication recommendation combines patient medical history with biomedical knowledge to assist doctors in determining medication combinations more accurately and safely. Existing works based on molecular knowledge neglect the 3D geometric structure of molecules and fail to learn the high-dimensional information of medications, leading to structural confusion. Additionally, it does not extract key substructures from a single patient visit, resulting in the failure to identify medication molecules suitable for the current patient visit. To address the above limitations, we propose a bimodal molecular recommendation framework named BiMoRec, which introduces 3D molecular structures to obtain atomic 3D coordinates and edge indices, overcoming the inherent lack of high-dimensional molecular information in 2D molecular structures. To retain the fast training and prediction efficiency of the recommendation system, we use bimodal graph contrastive pretraining to maximize the mutual information between the two molecular modalities, achieving the fusion of 2D and 3D molecular graphs. Additionally, we designed a molecular multi-step enhancement mechanism to re-calibrate the molecular weights.
Specifically, we employ a pre-training method that captures both 2D and 3D molecular structure representations, along with substructure representations, and leverages contrastive learning to extract mutual information. We then use the pre-trained encoder to generate molecular representations, enhancing them through a three-step process: intra-visit, molecular per-visit, and latest-visit. Finally, we apply temporal information aggregation to generate the final medication combinations. Our implementation on the MIMIC-III and MIMIC-IV datasets demonstrates that our method achieves state-of-the-art performance. Compared to the second-best baseline, our model improves accuracy by 0.61\%, while maintaining the same level of DDI and model efficiency. Our source code is publicly available at: \href{https://github.com/guangyunms/BiMoRec}{https://github.com/guangyunms/BiMoRec}.
\end{abstract}


\begin{keywords}
 Intelligent healthcare management\sep 
 Medication recommendation\sep 
 Recommender systems
\end{keywords}

\maketitle

\section{Introduction}
Medication recommendation is a crucial field in modern medical informatics, focused on providing personalized medication suggestions to doctors and patients by analyzing personal health data such as electronic health records (EHRs) and medical literature. With the rapid advancements in big data, artificial intelligence, and machine learning, smart healthcare has become an essential approach to addressing future challenges in medical resource distribution. Unlike traditional medication recommendations that rely mainly on doctors' experience and guidelines, modern systems can utilize extensive clinical datasets and literature databases. This allows for the automatic extraction of useful information and intelligent analysis, thereby enhancing the accuracy and efficiency of recommendations.

Early medication recommendations~\citep{earlywork1,earlywork2,earlywork3} primarily focus on a patient's most recent visit records, recommending medications based on the patient's current condition. Later research incorporates the patient's longitudinal historical visit records~\citep{longitudinal1,longitudinal2,longitudinal3}, thereby improving the accuracy of the recommendations.

However, unlike other fields such as e-commerce, which focus solely on the accuracy of recommendations, medication recommendations must consider not only accuracy but also safety. This is because different medications can interact, potentially causing adverse effects and impacting the patient's health. For example, both aspirin and warfarin are anticoagulants, but their simultaneous use may increase the risk of bleeding. A study~\citep{bhoi2021personalizing} shows that in the largest publicly available benchmark dataset, MIMIC-III, patients have an average of 20 different medications. This highlights the significant challenge of eliminating all potential drug-drug~\footnote{The terms "drug" and "medication" are used interchangeably in this paper.} interactions (DDIs)~\citep{ddi1,ddi2,ddi3}. Therefore, there is an urgent need for a methodology in medication recommendations that addresses both accuracy and safety concurrently.

To balance accuracy and safety, researchers explore the relationships between medical entities from EHRs. COGNet~\citep{cognet} retrieves a patient's historical diagnoses and medication recommendations and examines their relationship with current diagnoses to suggest medications. SafeDrug~\citep{safedrug} takes this further by utilizing the molecular structures of medications, encoding the connectivity and functionality of medication molecules, and explicitly modeling DDIs. A recent study, MoleRec~\citep{molerec}, proposes a molecular substructure-aware encoding method that models the interactions between substructures to identify those that truly contribute to patient healing. These existing methods primarily model medications from a coarse-grained perspective (medication entities) and a fine-grained perspective (molecular structures of medications). In comparison, fine-grained methods are superior to coarse-grained methods because they can uncover more detailed and complex relationships that are difficult to achieve with coarse-grained methods. Specially, these fine-grained studies~\citep{safedrug,molerec} improve the performance of medication recommendations by representing medications as finer-grained 2D molecular substructures, which helps model the pharmacological properties and DDIs from a more microscopic perspective. While they achieve a certain balance between accuracy and safety, they still have limitations.

\begin{figure}
    \centering
    \includegraphics[width=0.8\linewidth]{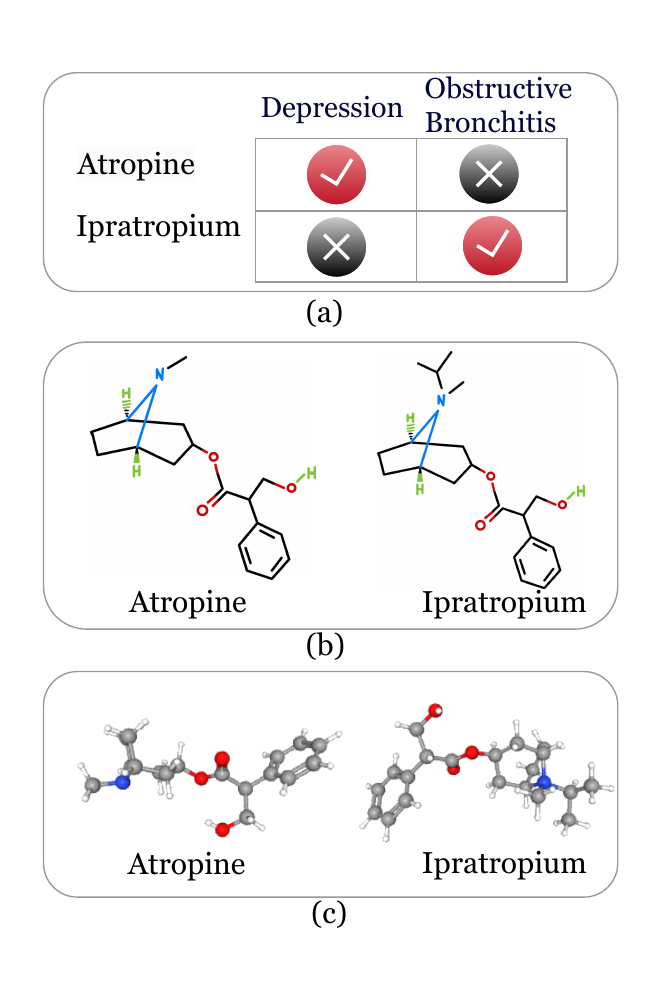}
    \vspace{-0.3in}
    \caption{Spatial structure representation of Atropine and Ipratropium. (a) Atropine is a medication for treating depression, while Ipratropium is used for treating obstructive bronchitis; (b) 2D molecular structure representation; (c) 3D molecular structure representation.}
    \label{fig:2d_3d}
    \vspace{-0.3in}
\end{figure}

Previous studies primarily focus on the 2D molecular structures of medications, overlooking their 3D spatial structures at the molecular level. This oversight significantly impacts the accuracy and safety of medication recommendations. For instance, as shown in Figure~\ref{fig:2d_3d} (a), Atropine treats depression, while Ipratropium addresses obstructive bronchitis. Using PubChem\footnote{\url{https://pubchem.ncbi.nlm.nih.gov/}}, a public database, we display the distinct structural representations of these two medications. Figure~\ref{fig:2d_3d} (b) illustrates that previous approaches, which only consider 2D molecular structures, show Atropine and Ipratropium as having similar 2D structures. This similarity can lead to erroneous recommendations, such as prescribing Ipratropium for depression patients or Atropine for those with obstructive bronchitis. These inaccuracies not only compromise recommendation precision but also pose significant health risks to patients. However, by analyzing the 3D spatial structures depicted in Figure~\ref{fig:2d_3d} (c), we find notable differences despite their similar 2D structures, with a Tanimoto coefficient \citep{rdkit} similarity of over 95\%. These differences indicate that 3D representation can effectively distinguish stereoisomers, such as enantiomers, which typically appear identical in 2D structures. The following advantages of 3D spatial structures are identified:

\textbf{Enhanced Recommendation Accuracy.} The 3D spatial structure resolves the issue of impaired recommendation performance due to the similarity of 2D structures, avoiding the noise interference caused by 2D representations. Additionally, 3D spatial structures provide a more accurate depiction of a molecule's true shape and spatial arrangement, aiding in the study of different pharmacological effects and assisting the recommendation system in making informed decisions.

\textbf{Alignment with Actual Medication Structures.} 3D spatial structures better represent the actual state of medications, as they inherently possess three-dimensional configurations. Traditional 2D molecular structures are often fragmented into smaller substructures during modeling. However, these substructures disrupt the medication's original spatial structure, resulting in a loss of integrity and stability.

\begin{figure}
    \centering
    \includegraphics[width=0.9\linewidth]{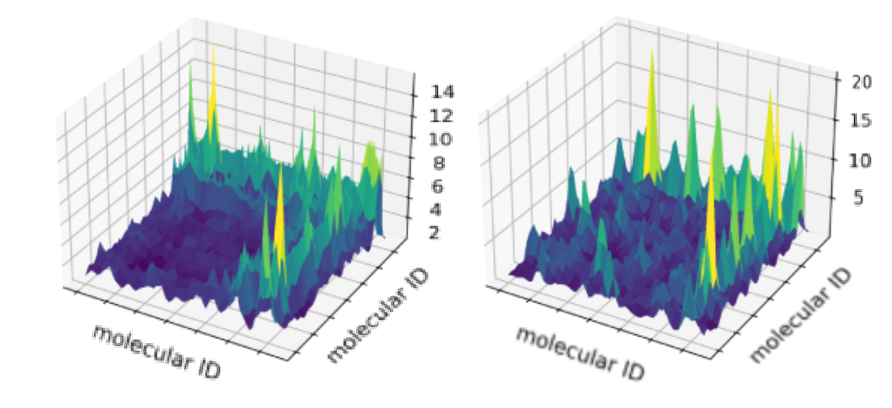}
    \caption{Comparison of molecular embedding similarity and molecular fingerprint similarity, with lighter colors indicating greater differences in similarity. The left side uses only the 2D modality, while the right side uses 2D and 3D modalities.}
    \label{fig:mol_heatmap_contract}
    \vspace{-0.3in}
\end{figure}

As shown in Figure~\ref{fig:mol_heatmap_contract}, we compute the similarity matrix between molecular embeddings and molecular fingerprints, subtract the matrices, and take the absolute values. Then, after data sampling and non-linear scaling, we plot a three-dimensional terrain map. Molecular fingerprints \citep{rdkit} are a digital representation that encodes the complex structural information of molecules as a series of bit vectors, commonly used for molecular identification. Our goal is to make molecular embeddings more distinguishable than molecular fingerprints. The left figure illustrates that the molecular embeddings do not differentiate significantly from the 2D molecular fingerprints in terms of similarity, with the overall terrain appearing uniform and evenly distributed, showing clear distinctions between high and low regions. Most molecular embeddings show no significant difference from their molecular fingerprints, which may lead to issues of structural confusion. In contrast, the right figure, with the inclusion of the 3D modality, demonstrates a more reasonable distribution of vectors in the embedding space. The terrain is notably more uneven, improving the distinguishability between molecular embeddings and fingerprints and optimizing the molecular embedding representation space. This experiment further demonstrates the advantages and significance of the 3D molecular modality in medication recommendation systems.

Although 2D structures are simpler in medication recommendation, they offer higher training efficiency compared to more complex 3D structures. Therefore, this paper continues to use 2D structures as in previous work, while incorporating the 3D concept proposed in this study, leading to the development of a bimodal molecular recommendation framework named BiMoRec. This model simultaneously addresses the confusion of 2D molecular structures and the inability of substructures to represent true molecular states, while also alleviating the issue of model training efficiency caused by the complexity of 3D structures.

First, the 2D modality extracts embeddings of the medication molecular structures and their substructures from molecular graphs using the GIN network \citep{gin}. Then, based on the medication molecule's SDF file, the corresponding 3D data is retrieved, and 3D descriptions of the substructures are generated. Using a 3D geometric vector machine \citep{gvp, luo2023calibrated}, high-dimensional embeddings of the molecular structures and their substructures are obtained. Finally, based on the embeddings from both the 2D and 3D modalities, a dual-modal graph contrastive learning module is designed to compare the mutual information between the two modality graphs \citep{ntxent}, enabling joint learning of medication molecular representations.

Then, this paper employs a multi-step molecular enhancement mechanism to learn the correlation between the patient's health conditions and the molecular dual-granularity structure. First, self-attention mechanisms are applied to the medication molecules in each visit, improving their interactions and enabling the recommendation algorithm to identify medications that may be effective for the disease. Next, the relationships between molecules and diseases, medications, and procedures are analyzed to enhance the model’s learning ability. Finally, based on the patient's current condition, relevant molecules are identified, and suitable medications are recommended.

Specifically, the main contributions of this paper can be summarized as follows:

\begin{itemize}
\item We, for the first time, incorporate the 3D spatial structure information of medications to differentiate their representations from a more microscopic view, helping the model better distinguish important features from noise, reduce bias and enhance overall performance.
\item We simultaneously utilize both the 2D and 3D modalities of medication molecules, along with a dual-modal graph contrastive learning module, to help the model capture key details that may be missed by a single modality. This simultaneous use provides a more comprehensive representation of molecular features, leading to a deeper understanding of molecular properties and stronger generalization capabilities for the model.
\item We employed a multi-step molecular enhancement mechanism to assist the model at various levels, enabling more effective recommendations and improving both the accuracy and safety of the recommendations.
\item We conduct extensive experiments on two real datasets, MIMIC-III and MIMIC-IV, and the results demonstrate that our proposed method outperforms existing methods across multiple metrics.
\end{itemize}

The structure of this paper is outlined as follows: (1) Introduction: Highlights the key innovations and the motivation for this study. (2) Related Work: Reviews significant studies and current trends in medication recommendation research. (3) Problem Definition: Defines the relevant medical concepts, entities, and the inputs and outputs used. (4) Methods: Explains the main ideas of the model and its implementation details. (5) Experiments: Describes the experimental setup, benchmarks, evaluation methods, and results. (6) Discussion: Analyzes the experimental outcomes and introduces a series of supplementary experiments. (7) Conclusion: Summarizes the key findings and suggests directions for future research.

\section{Related Works}
In this section, we review related work in two key areas: medication recommendation and medication molecular representation.

\subsection{Medication Recommendation}
In recent years, the field of medication recommendation has evolved through several stages, with the goal of offering personalized medication plans for patients. Early medication recommendation methods, such as instance-based approaches~\citep{leap}, focused on recommending medications based on single-visit patient data. Subsequent studies introduced sequential models~\citep{retain} to account for longitudinal data from patients' historical visit records.


Existing medication recommendation tasks can be divided into three categories:
(1) The first approach views medication recommendation as a conventional recommendation problem, focusing on the underlying techniques in the recommendation domain. RETAIN \citep{retain}, based on a two-level attention model, identifies key time points in patients' historical visits. GAMENet \citep{gamenet} enhances medication systems by leveraging graph neural networks. COGNet \citep{cognet} employs transformers for medication recommendation, adopting a translation-based approach to recommend medications from medical entities. Another study \citep{jin2018treatment} employs three LSTMs to capture relationships between heterogeneous data, predicting necessary medication combinations.
(2) The second approach emphasizes the relationships between medical entities, including diseases, procedures, and medications. DPR \citep{drug_package} represents interactions between medical entities as a graph and designs a medication recommendation module based on graph neural networks to capture correlations among entities. DPG \citep{drug_package_generation} takes into account drug-drug interactions (DDI), constructing a medication interaction graph based on patient history and related knowledge. StratMed \citep{StratMed} employs a dual-graph architecture to balance accuracy and safety, collaboratively enhancing model performance.
(3) The third approach leverages external knowledge to enhance the relationships among medical entities, improving model performance. Carmen \citep{carmen} integrates patient records into molecular representation learning to differentiate between molecules. SafeDrug \citep{safedrug} introduces detailed molecular information to enhance medication representations, aiming to reduce the risk of DDI and promote safer medication combinations. MoleRec \citep{molerec} further builds relationships between molecules and substructures to improve medication recommendations, enabling more reliable predictions of potential medication interactions and uncovering more comprehensive medication information.

\subsection{Molecular Representation Learning}

In existing studies that enhance medication recommendation performance using external knowledge, the majority rely on molecular knowledge of medications.

Carmen \citep{carmen} fuses patient records with molecular representations to further differentiate molecules, enhancing recommendation outcomes. SafeDrug \citep{safedrug} constructs 2D molecular modality data into graph structures and employs GNN to capture drug-drug interaction (DDI) relationships between molecules, thereby improving the safety of medication recommendations. MoleRec \citep{molerec} not only utilizes 2D molecular data but also segments molecules into multiple substructures, building mappings between molecules and substructures to uncover deeper medication relationships. MEGACare \citep{megacare} integrates 2D molecular data into a multi-perspective hypergraph prediction framework, leveraging molecular knowledge to enhance the representation of medical codes.

In research utilizing external knowledge, integrating molecular knowledge has become a reliable approach to improve model performance. However, previous work only considers the 2D modality of molecules, resulting in molecular representations lacking high-dimensional information. This leads to structural confusion and feature overlap among molecules, making it difficult to clearly distinguish similar molecules in the representation space, thereby reducing the overall discriminative power of the representation space.

Researchers have long recognized the importance of 3D geometric information in the molecular domain. Initially, 3D molecular structures were introduced to predict the binding sites and affinity between medications and targets \citep{3d_good_1}, solving practical challenges associated with 3D structure applications. Subsequently, studies explored the role of 3D quantitative structure-activity relationship (3D-QSAR) \citep{3d_good_2} models in medication design, providing a foundation for medication prediction and optimization. KDBNet \citep{luo2023calibrated} utilizes 3D protein and molecular structures to predict the binding affinity between protein kinases and compounds, incorporating 3D properties often neglected by other models. Recent review articles \citep{3d_good_3} have overviewed the latest applications of 3D geometric structures in bioorganic and medicinal chemistry, highlighting their potential in structure-based medication discovery and design.

Multi-modal refers to the integration of multiple types of data or information from different sources or modalities. Molecular multimodal refers to the incorporation of various types of molecular data, such as 2D structural information and 3D geometric information, to enhance molecular representation and analysis. In the field of multi-modal molecular learning, 2D graph structures and 3D geometric structures can be viewed as different perspectives of the same molecule. Inspired by contrastive pre-training methods in the vision domain, many research efforts have begun to explore how to use 2D and 3D information jointly for molecular pre-training. For example, using two encoders to separately encode 2D and 3D molecular information while maximizing the mutual information between their representations \citep{Multi_Modal_1}. Alternatively, pre-training 2D and 3D encoders through contrastive learning and reconstruction \citep{Multi_Modal_2}. Recent studies have unified the aforementioned 2D and 3D pre-training methods and proposed a 2D graph neural network model \citep{Multi_Modal_3} that can be enhanced by 3D geometric features.


Our research aims to integrate molecular multi-modality without compromising the efficiency and complexity of the recommendation system, in order to address the issue of structural confusion in medication recommendation and significantly improve existing models.

\section{Problem Definition}

\subsection{Medical Entity}
In this study, medical entities primarily include three types: diseases, procedures, and medications, denoted as \(\mathcal{D} = \{d_1, d_2, \ldots\}\), \(\mathcal{P} = \{p_1, p_2, \ldots\}\), and \(\mathcal{M} = \{m_1, m_2, \ldots\}\). Additionally, the molecular data come from the PubChem open-source database.

\subsection{Input and Output}
This paper uses EHR as the data source, containing comprehensive records of patient visits and treatments. Each patient record is represented as \(\mathcal{H}\), which consists of multiple visit entries \(\mathcal{H} = \{v_1, v_2, \ldots, v_t\}\), where \(v_t\) corresponds to the clinical information for the $t\text{-th}$ visit. Each visit \(v_t = \{\mathcal{D}_t, \mathcal{P}_t, \mathcal{M}_t\}\) includes three elements: \(\mathcal{D}_t \subset \mathcal{D}\), \(\mathcal{P}_t \subset \mathcal{P}\), and \(\mathcal{M}_t \subset \mathcal{M}\), representing the patient's medical data for both historical and current visits. The model's output is denoted as \(\hat{\mathcal{M}}_t\), which is the predicted medication combination for the visit \(v_t\).

\subsection{DDI Matrix}

DDIs are a critical concern in medication recommendations, as they highlight combinations of medications that may present confirmed safety risks to patients. To enhance the safety of medication therapy, it is essential to exclude combinations that carry such risks, thereby minimizing the occurrence of adverse events. 

We obtain our DDI data from the Adverse Event Reporting Systems~\citep{AERS}. This information is represented in a binary matrix \({\mathbf{M}^{ddi}} \in \{0,1\}^{|\mathcal{M}| \times |\mathcal{M}|} \), where an entry of \(\mathbf{M}^{ddi}_{ij} = 1\) denotes an interaction between medications \(m_i\) and \(m_j\). A high occurrence of DDIs in this matrix indicates potential safety concerns in the recommended medication combinations. For example, in the MIMIC-III dataset \citep{mimic3}, the actual rate of DDIs is approximately 6\%~\citep{safedrug}. This indicates that a considerable number of medication combinations may pose safety risks to patients, emphasizing that drug-drug interactions are a significant concern in real-world clinical settings. Special attention is needed to avoid recommending combinations that could lead to adverse reactions.

\section{Methods}

\begin{figure*}
    \centering
    \includegraphics[width=\textwidth]{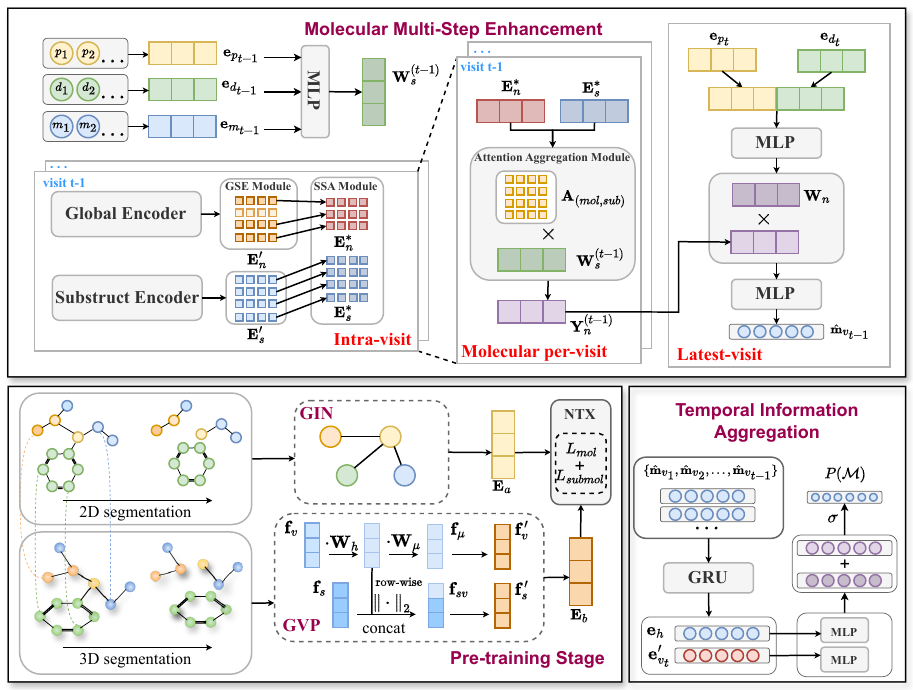}
    \caption{The pre-training phase illustrated in the bottom left side of the figure demonstrates the bimodal fusion process of molecular structure and decomposed substructures, and its results are applied to the downstream tasks shown in the top left side. The molecular multi-step enhancement phase in the upper corner explains how molecular knowledge interacts with patient representations at the visit level. It sequentially performs molecular enhancement at three hierarchical levels: intra-visit, molecular per-visit, and latest-visit. The temporal information aggregation phase on the lower right side of the figure models the sequence relationships of the interaction results and maps them to recommendation probabilities, obtaining the final recommendation results.}
    \label{fig:model}
    \vspace{-0.3in}
\end{figure*}

Figure~\ref{fig:model} illustrates our model, consisting of three primary components.

(1) In the pre-training stage, we integrate dual-modality molecular information using contrastive learning. This allows us to obtain a molecular encoding module and a substructure encoding module that incorporate knowledge from both modalities. It can also be understood as the 3D modality teacher model transferring knowledge into the 2D model.

(2) In the molecular multi-step enhancement phase, our goal is to encode the molecules and their substructures using the encoder developed in the pre-training stage, generating embedding matrices. Through a multi-step enhancement mechanism, we apply triple enhancement of molecular representations at the visit level. This process increases the weight of key molecules to identify critical medications.

(3) In the temporal information aggregation stage, our objective is to integrate patient progression information derived from medication representations at different temporal points. By adjusting the probability bias of each medication based on the patient's current health condition, we can recommend medications that exceed a predefined threshold.

\subsection{Pre-training}

In this phase, our aim is to integrate the dual-modality information of molecules, which involves the representation of 2D and 3D structures as well as mutual information learning. To ensure the training efficiency of the model and reduce complexity, a contrastive learning pre-training method is employed.

For the 2D molecular modality, we first represent it as a graph structure. The medication molecule consists of atoms and bonds, where atoms serve as nodes and chemical bonds as edges. Given a molecule \( G \), the representation of the entire graph is generated through \( l \) layers of Graph Neural Networks (GNN).

In this paper, we use GIN \citep{gin}, a variant of GNN, as the encoder for medication molecular graphs to obtain holistic-level representations, which is commonly used in recent work related to medication recommendation and molecular research \citep{mole_relate_work_1, mole_relate_work_2, mole_relate_work_3}. GIN iteratively updates node features by aggregating features of neighboring nodes, i.e., updating representations of atomic nodes through aggregation strategies. In general, we have:
\begin{gather}
    \mathbf{f}_{a}^{(l)}=\mathrm{UPDATE}\left(\mathbf{f}_{a}^{(l-1)},\mathbf{f}_{b}^{(l-1)},\mathbf{e}_{(a,b)}|b\in\mathcal{N}(a)\right), \\
    \mathbf{E}_g=\text{READOUT}\left(\left\{\mathbf{f}_{a}^{(l)}|a\in V\right\}\right),
\end{gather}
where \(\mathbf{f}_{a}^{(l)}\) is the node feature of node \(a\) in the layer \(l\), \(\mathbf{e}_{(a,b)}\) is the edge feature of edge \((a,b)\), and \(\mathcal{N}(a)\) denotes the neighbor of node \(a\) on the graph. \(\mathrm{UPDATE}\) function updates the feature representation of a node by aggregating its own features and neighboring features via weighted summation.
Subsequently, \(\mathrm{READOUT}\) function is employed to aggregate all the node features from the \( l\text{-th} \) layer.

Through the aforementioned process, the representations of all medication molecules are obtained, forming an embedding matrix \(\mathbf{E}_{n}\in \mathcal{R}^{|\mathcal{M}|\times dim}\), where each row represents a medication molecule. This matrix will be used in the subsequent contrastive learning phase.

For the 3D molecular modality, given the 3D coordinates of an atom, we represent the molecular structure as a graph where atoms serve as nodes. We define an edge between two atoms if the distance between them is less than 4.5 Å \citep{atom3d}. The symbol Å stands for angstrom, a unit of length commonly used in chemistry, physics, and biology to measure very small distances, such as atomic and molecular scales. In the constructed 3D molecular graph, each node and edge is linked to geometric features like 3D coordinates and scalar features that describe chemical properties. Thus, we employ Geometric Vector Perceptrons (GVPs) to encode the molecular structure. Its main advantage lies in its design specifically considering 3D data, enabling direct learning of structural representations from raw atomic coordinates without the need to construct features that are invariant to rotation and translation \citep{gvp}.

In actual model construction, we use the tuple \( (\mathbf{f}_s , \mathbf{f}_v) \) to represent the scalar features and vector features of atom independently. The molecular GVP transforms node and edge features through graph convolution layers to obtain the representation of the input molecule. 
The specific process of feature learning and transformation is as follows:
\begin{gather}
    \mathbf{f}_{sv} = \mathrm{CONCAT}({\|\mathbf{W}_h\mathbf{f}_v\|}_{2, \text{row-wise}}, \mathbf{f}_s), \\
    \mathbf{f}_s^{\prime} = \sigma(\mathbf{W}_v\mathbf{f}_{sv}+b), \\
    \mathbf{f}_\mu = {\|\mathbf{W}_\mu\mathbf{W}_h\mathbf{f}_v\|}_{2, \text{row-wise}}, \\
    \mathbf{f}_v^{\prime} = \sigma(\mathbf{f}_\mu)\odot\mathbf{f}_\mu,
\end{gather}
where \( W_h \), \( W_v \), and \( W_\mu \) are three weight matrices for transformation, \( b \) is a bias term, \( \odot \) represents element-wise multiplication, \(\sigma\) is a nonlinear function, and \(\mathrm{CONCAT}\) concatenates the row-wise L2 norm with \(\mathbf{f}_s\).
After passing through the GVP graph convolution layer, \( (\mathbf{f}_s, \mathbf{f}_v) \) is transformed into \( (\mathbf{f}_s^{\prime}, \mathbf{f}_v^{\prime}) \). Each node aggregates embeddings from its neighboring nodes and edges to update its own representation.


After the final layer of molecular GVP, a global additive pooling operation is applied to aggregate all node representations into the scalar representation of the input medication, obtaining the 3D embedding matrix \( \mathbf{E}_{nt} \in \mathcal{R}^{|\mathcal{M}|\times dim} \) of the molecule, where each row represents a molecule's representation.

For molecular substructures, existing work demonstrates that the properties of medication molecules are partially determined by specific substructures, and the treatment of a patient's disease also depends on the functionality of certain substructures. Our goal is to reconstruct the 3D structures of substructures to bring them closer to the actual molecular structures. We first use the BRICS \citep{molerec} method to decompose medication molecules into substructures, and then complete the 3D structure of the substructures using the API provided by RDKit \citep{rdkit}. We decompose all medication molecules to obtain a substructure set \(S\), and finally obtain the embedding matrices of the two different modalities of the substructures \(\mathbf{E}_{s}, \mathbf{E}_{st} \in \mathcal{R}^{|S|\times dim}\), where each row represents the embedding of a specific substructure.

In the subsequent process, contrastive learning achieves two objectives. The first objective is to make molecules with the same index \( i \) more similar. The second objective is to enforce differences between negative samples \( \mathbf{E}_{a_i} \) and \( \mathbf{E}_{b_k} \), where \( i \neq k \), i and k represent the \( i\text{-th} \) embedding in the 2D embedding table and the \( k\text{-th} \) embedding in the 3D embedding table. We employ the commonly used NTXent loss \citep{luo2023calibrated, ntxent} to jointly optimize our model:

\begin{gather}
\mathrm{L}(\mathrm{\mathbf{E}_a,\mathbf{E}_b})=-\frac{1}{N}\sum_{i=1}^{N}\left[\log\frac{e^{sim(\mathbf{E}_{a_i},\mathbf{E}_{b_i})/\tau}}{\sum_{k\neq i}^{N}e^{sim(\mathbf{E}_{a_i},\mathbf{E}_{b_k})/\tau}}\right], \\
sim(\mathbf{E}_{a}, \mathbf{E}_{b}) = \mathbf{E}_{a} \cdot \mathbf{E}_{b}/(\|\mathbf{E}_{a}\|\|\mathbf{E}_{b}\|),
\end{gather}
where \(sim(\cdot)\) is the cosine similarity and \(\tau\) is a temperature parameter which acts as a weight for the most similar negative pair.

We apply this loss function to molecules and their substructures by inputting \(\mathbf{E}_{n}\) and \(\mathbf{E}_{nt}\) into the aforementioned formula to obtain \(\mathcal{L}_{mol}\), and inputting \(\mathbf{E}_{s}\) and \(\mathbf{E}_{st}\) into the same formula to obtain \(\mathcal{L}_{submol}\). Finally, we combine them to obtain the total loss function.
\begin{gather}
    \mathcal{L}_{mol} = \mathrm{L}(\mathbf{E}_{n}, \mathbf{E}_{nt}), \\ \mathcal{L}_{submol} = \mathrm{L}(\mathbf{E}_{s}, \mathbf{E}_{st}), \\
    \mathcal{L}_{total} = \mathcal{L}_{mol}+\mathcal{L}_{submol}.
\end{gather}

\subsection{Multi-Step Molecular Enhancement}
At this stage, we will enhance the medication molecules at three levels: intra-visit, molecular per-visit, and latest-visit, to obtain the key molecular representations required by the patients.

We encode the diseases, procedures, and medications as part of the patient's current health environment. We first extract the sets of diseases, procedures, and medications at each time point from the visit \( v_{t-1} \). We define three learnable embedding matrices, \( \mathbf{E}_d \), \( \mathbf{E}_p \), and \( \mathbf{E}_m \), corresponding to diseases, procedures, and medications. We generate \( \mathbf{e}_{d_i}\), \( \mathbf{e}_{p_i} \), and \( \mathbf{e}_{m_i} \) by mapping diseases \( d_i \), procedures \( p_i \), and medications \( m_i \) to the embedding space, and use them as entity-level representations shared globally for downstream tasks. We obtain the set representations \( \mathbf{e}_{d_{t-1}} \), \( \mathbf{e}_{p_{t-1}} \), and \( \mathbf{e}_{m_{t-1}} \) by mapping all diseases, procedures, and medications through their corresponding embedding matrices.
\begin{gather}
    \mathbf{e}_{d_{t-1}}=d_{t-1}\mathbf{E}_d, \quad \mathbf{e}_{p_{t-1}}=p_{t-1}\mathbf{E}_p, \quad \mathbf{e}_{m_{t-1}}=m_{t-1}\mathbf{E}_m.
\end{gather}

Finally, we concatenate the three vectors of equal length, resulting in a single vector with three times the original length, forming the visit-level representation for the visit at time \( t-1 \).
\begin{equation}
    \mathbf{e}_{v_{t-1}}^{\prime}=\mathrm{CONCAT}(\mathbf{e}_{d_{t-1}} \| \mathbf{e}_{p_{t-1}} \| \mathbf{e}_{m_{t-1}}).
\end{equation}

Then, a multi-layer perceptron (MLP) converts the dimension of the vector \( \mathbf{e}_{v_{t-1}}^{\prime} \) to the total number of molecular substructures, and a sigmoid function modulates the relevance of each substructure at time \( t-1 \).
\begin{equation}
    \mathbf{W}^{(t-1)}_{s}=\sigma(\mathrm{MLP}(\mathbf{e}_{v_{t-1}}^{\prime})).
\end{equation}

Following the initialization process outlined above, we will now provide a detailed description of our molecular multi-enhancement mechanism.

\subsubsection{Intra-visit Enhancement}

After completing the pre-training process in the previous stage, two encoders that integrate dual molecular modalities were derived: a global encoder and a substructure encoder, for use in subsequent molecular representation tasks.

We employ the global encoder and substructure encoder trained during the pretraining phase to encode molecular structures and their substructures, obtaining \( \mathbf{E}'_n \) and \( \mathbf{E}'_s \). For substructures, as the function of molecules is realized through a set of substructures, it is necessary to model the interactions between these substructures. Therefore, we introduce a Substructure Self-Attention (SSA) module \citep{ssa} to achieve this goal. We use the embedding matrix \( \mathbf{E}'_s \) of all substructures as input to establish an attention mechanism between each substructure, and finally output an embedding matrix of equal size \( \mathbf{E}^*_s \in \mathbb{R}^{|\mathcal{S}| \times dim} \). The self-attention layer is defined as follows:
\begin{gather}
    \mathrm{H=LayerNorm(\mathbf{E}'_s+\mathbf{A}_{self}(\mathbf{E}'_s)),}\\
    \mathrm{\mathbf{E}^*_s=SSA(\mathbf{E}'_s)=LayerNorm(H+MLP(H))}.
\end{gather}

\( \mathcal{Attn}_{self} \) denotes the self-attention mechanism \citep{attention} defined by the following formula:
\begin{equation}\mathbf{A}_{self}(\mathrm{X})=\mathrm{Softmax}\left(\frac{\mathbf{Q}\mathbf{K}^\top}{\sqrt{d}}\right)\mathbf{V},\end{equation}
where \( d \) represents the dimensionality of the key embeddings, and \( \mathbf{Q} \), \( \mathbf{K} \), and \( \mathbf{V} \) are the query, key, and value matrices derived from the input \( \mathbf{X} \).

For molecular structures, we need to select from the embedding table \(\mathbf{E}'_n\) based on the patient's visit data and perform molecular enhancement. We input \(\mathbf{E}'_n\) into the global selection enhancement module (GSE) to obtain the enhanced result \(\mathbf{E}^*_n \in \mathbb{R}^{|\mathcal{M}| \times dim}\):
\begin{equation}
\mathbf{E}^*_n = \text{GSE}(\text{SSA}(\mathbf{E}'_{n} )).
\end{equation}

Specifically, the GSE module searches for the corresponding embedding vectors based on the medication molecule combinations relevant to the current context, constructs a small embedding matrix, and feeds it into the attention module. This process generates an enhanced matrix, which is then used to update the global molecular embedding matrix.

\subsubsection{Molecular per-visit Enhancement}

At this stage, we will perform molecular enhancement based on each visit. Specifically, we input the molecular-level representation and substructure interaction representation, \( \mathbf{E}^*_n \) and \( \mathbf{E}^*_s \), into the Attention Aggregation Module to perform attention-based aggregation. When calculating the attention coefficients, we use a mask matrix to adjust the attention matrix.
\begin{gather}
\mathbf{A}_{(mol, sub)}(\mathrm{X,Y})=\mathrm{mask}\left(\mathrm{Softmax}\left(\frac{\mathbf{Q}_x\mathbf{K}_y^\top}{\sqrt{d_{min}}}\right)\right),
\end{gather}
where \( d_{\min} \) is the minimum embedding dimension of \( \mathbf{X} \) and \( \mathbf{Y} \), and the mask function replaces the set of substructures that lack corresponding mappings with the minimum value.
\begin{equation}
\forall s_i\in S,\mathbf{y}_{m_i}=\mathbf{A}_{(mol, sub)}\mathbf{W}^{(t-1)}_{s}\mathbf{E}^*_{s_i},
\end{equation}
where \(\mathbf{E}^*_{s_i}\) is the vector of the corresponding substructure \(s_i\) in \(\mathbf{E}^*_s\). Then, we obtain the contextual representation of the substructure at the current time sequence through the attention matrix.

Ultimately, we derive all the aggregated molecular embedding representations \(\mathbf{Y}^{(t-1)}_{n}\) for the visit at time \( t-1 \).

\subsubsection{Latest-visit Enhancement}

To relate the aggregated molecular embedding to the patient's latest health condition, we use the most recent patient visit as a reference for final enhanced. Through an MLP layer, we obtain the weights of the latest patient representation for the elements of the molecular set to calibrate the molecular embedding accordingly.
\begin{gather}
    \mathbf{e}_{d_t}=d_t\mathbf{E}_d, \quad \mathbf{e}_{p_t}=p_t\mathbf{E}_p,\\
    \mathbf{e}_{v_t}^{\prime}=\mathrm{CONCAT}(\mathbf{e}_{d_t} \| \mathbf{e}_{p_t}),\\
    \mathbf{W}_{n}=\sigma(\mathrm{MLP}(\mathbf{e}_{v_t}^{\prime})).
\end{gather}

\( \mathbf{e}_{d_t} \) and \( \mathbf{e}_{p_t} \) are the embeddings of the diagnosis and procedure representations from the patient's current visit. By concatenating these embeddings and passing them through an MLP layer, we obtain the weight vector \( \mathbf{W}_{n} \). Finally, we correct the molecular embedding matrix by multiplying it with a diagonal matrix, and then adjust the dimensions through an MLP network to obtain the representation \( \hat{\mathbf{m}} \) for each medication.
\begin{equation}
\hat{\mathbf{m}}_{v_{t-1}}=\mathrm{MLP}(\mathbf{W}_{n}\mathrm{Y}^{(t-1)}_{n}).
\end{equation}

By enhancing the molecular representations based on recent patient visits, the most recent patient conditions were incorporated into the molecular matrix for each visit.

\subsection{Temporal Information Aggregation}
The purpose of this stage is to integrate the medication representations generated for the patient at each time point. This representation is derived from multiple dynamic enhancements of molecules, ultimately yielding the recommendation probabilities for all medications.

By embedding the multiple medication representations generated in the previous stage \(\{\hat{\mathbf{m}}_{v_{1}}, \hat{\mathbf{m}}_{v_2}, \ldots, \hat{\mathbf{m}}_{v_{t-1}}\}\) and inputting them into a gated recurrent unit \( \text{GRU}(\cdot) \) to model the sequential relationships, \( \{v_1, v_2, \ldots, v_{t-1}\} \) are patient visits \(t-1\) times, we then process the outputs through a multilayer perceptron.
Finally, we concatenate the results with the patient's most recent representation $\mathbf{e}_{v_t}^{\prime}$ to generate the final medication probabilities.
\begin{gather}
    \mathbf{e}_h = \text{GRU}(\hat{\mathbf{m}}_{v_{1}}, \hat{\mathbf{m}}_{v_2}, \ldots, \hat{\mathbf{m}}_{v_{t-1}}),\\
    P(\mathcal{M}) = \sigma(\mathrm{MLP}(\mathbf{e}_h)+\mathrm{MLP}(\mathbf{e}_{v_t}^{\prime})),
\end{gather}

Using the nonlinear activation function $\sigma$, the input is converted into the predicted probability $P(\mathcal{M})$ for each medication. If the probability exceeds a certain threshold, the recommendation for this medication is accepted; otherwise, the recommendation is rejected. Ultimately, this process yields the predicted medication combination \(\hat{\mathcal{M}}_t\) for the current visit \( v_t \).

\subsection{Model Training}
Medication recommendation can be framed as a multi-label binary classification task. Thus, we use the binary cross-entropy loss function \( L_{bce} \) and the multi-label margin loss function \( L_{multi} \). Additionally, given the unique nature of medication recommendation, we use the medication DDI loss function \( L_{ddi} \) \citep{molerec}, which calculates the probability of DDI occurrence by identifying medication pairs with potential DDI risk in the medication combination. The definitions of the three loss functions can be defined as:
\begin{gather}
    \mathcal{L}_{bce} = -\sum_{i=1}^{|\mathcal{M}|}{m_i}\log({\hat{m}_i})+(1-{m_i})\log (1-{\hat{m}_i}),\\
    \mathcal{L}_{multi} = \sum_{i,j:{m_i}=1,{m_j}=0} \frac{\max(0,1-({\hat{m}_i}-{\hat{m}_j}))}{|\mathcal{M}|},\\
    \mathcal{L}_{ddi} = \sum_{i=1}^{|\mathcal{M}|}\sum_{j=1}^{|\mathcal{M}|} \mathbf{M}^{ddi}_{ij}\cdot{\hat{m}_i}\cdot {\hat{m}_j},
\end{gather}    
where \(m_i\) represents the true value of the \(i\text{-th}\) medication during the current visit, while \(\hat{m}_i\) represents the model's predicted value for the \(i\text{-th}\) medication, which are both binary variables.

Although DDI reflects the potential risk of medication combinations, from a clinical perspective, certain adverse reactions are acceptable. Pursuing the lowest DDI rate without considering efficacy may not result in the optimal medication combination. Therefore, we must set a threshold for DDI; if this threshold is surpassed, the weight assigned to DDI increases. When using the composite loss function, we adopt the same research methodology as before \citep{molerec, loss_balance} to achieve a balance between the accuracy and safety of the medication combinations.
\begin{gather}
    \mathcal{L} = \alpha (\beta \mathcal{L}_{bce}+(1-\beta)\mathcal{L}_{multi})+(1-\alpha )\mathcal{L}_{ddi},\\
    \alpha=
    \begin{cases}
    1  &\rho<\phi  \\
    \min\{1,\exp(\tau(1-\frac\rho\phi))\}  &\rho\geq\phi
    \end{cases},
\end{gather}
where $\beta$ is hyperparameters, and the controllable factor $\alpha$ is relative to DDI rate, $\phi \in (0,1)$ is a DDI acceptance rate, $\rho$ is the current DDI rate and $\tau$ is a hyper-parameter.

\section{Experiments}
This section provides an overview of the datasets used and a comparison of the results from the baseline models and our model. Additionally, we present an explanation of the evaluation metrics and the specific experimental setup, including model configuration and parameter selection.

\subsection{Datasets}

\begin{table}
    \caption{Statistics of the datasets.}
    \begin{tabular}{|c|c|c|}
    \toprule
    Item & MIMIC-III & MIMIC-IV \\
    \midrule
    \# patients         & 6,350  & 60,125 \\
    \# visits events  & 15,032 & 156,810 \\
    \# diseases         & 1,958  & 2,000 \\
    \# procedures       & 1,430  & 1,500 \\
    \# medications      & 131   & 131\\
    avg. \# of visits    & 2.37  & 2.61\\
    avg. \# of medications & 11.44  & 6.66\\
    \bottomrule
    \end{tabular}
    \label{tab:datasets}
\end{table}

We utilize the MIMIC-III \citep{mimic3} and MIMIC-IV \citep{mimic4} datasets, which originate from clinical data in the intensive care unit (ICU). These datasets include clinical records, physiological measurements, laboratory monitoring data, and medication records of ICU patients. We follow the same data processing methods as previous studies. The detailed statistics of the processed datasets are presented in the table \ref{tab:datasets}.

Molecular SDF files are sourced from the PubChem database \citep{pubchem}, and the 3D data for substructures is generated using functions from the RDKit library.

\subsection{Baselines}
To verify the performance of our model, we chose the following high-performance methods as the baseline model for comparison.

\textbf{LR} (Logistic Regression) is a widely employed linear classification algorithm that calculates the probability of an instance belonging to a specific category based on a weighted combination of input features. It is commonly used for probability prediction and both binary and multi-class classification tasks.

\textbf{ECC} \citep{ecc} (Ensemble of Classifier Chains) utilizes a series of interconnected classifiers to improve prediction accuracy, where the output of one classifier serves as the input for the next. This method is well-suited to multi-label classification tasks and can effectively enhance the model's overall performance.

\textbf{RETAIN} \citep{retain} is an attention-based model that analyzes patient time series for precise disease prediction and management. It provides medication recommendations by adaptively identifying critical clinical events in a patient's medical history.

\textbf{LEAP} \citep{leap} optimizes medication efficacy by modeling label dependencies and decomposes the treatment process into a sequential procedure.


\textbf{GAMENet} \citep{gamenet} is a medication recommendation model that integrates the strengths of graph neural networks with memory networks and effectively discerns patterns and temporal sequences within medical data, thus improving the accuracy of its predictions.

\textbf{SafeDrug} \citep{safedrug} employs the combination of patients' health conditions and medication-related molecular knowledge. This approach, by reducing the impact of DDIs, offers safer medication combinations.

\textbf{MICRON} \citep{micron} recommends medications based on the dynamic changes in a patient's health condition. It updates medication combinations according to the patient's new symptoms to improve therapeutic outcomes while reducing potential side effects.

\textbf{COGNet} \citep{cognet} utilizes the Transformer architecture for medication recommendations, using a translation approach to derive medications from illnesses. It also features a copy mechanism to integrate beneficial medications from past prescriptions into new recommendations.

\textbf{MoleRec} \citep{molerec} investigates the importance of specific molecular substructures in medications. This approach enhances the precision of medication recommendations by utilizing finer molecular representations.

\textbf{CausalMed} \citep{causalmed} identifies causal relationships between medical entities through causal discovery, accounting for dynamic differences under varying health conditions and translating these into causally linked medication recommendations.

\subsection{Evaluation Metrics}

We employ four primary metrics, Jaccard, DDI rate, F1, and PRAUC, to assess the effectiveness of our approach. The role of each metric in our study is outlined below.

\textbf{Jaccard} (Jaccard similarity score) quantifies the similarity between two sets by dividing the size of their intersection by the size of their union. In medication recommendation, a higher Jaccard score indicates increased alignment between the predicted and actual medication combinations, indicating improved accuracy.

\textbf{DDI} (Drug-Drug Interaction rate) evaluates the incidence of adverse reactions in the recommended combinations. A lower rate indicates greater safety in the drug combinations.

\textbf{F1} (F1-score) combines precision and recall, which reflects the model's capacity to accurately detect correct medications while ensuring comprehensive coverage.

\textbf{PRAUC} (Precision-Recall Area Under Curve) evaluates model performance across different recall levels, demonstrating the capability to maintain precision with increasing recall.

\textbf{Avg.\# of Medications} (Average number of Medications)quantifies the average number of medications in each recommendation result. A higher value suggests that each recommendation includes more medications, which may increase the complexity of clinical treatment and the risk of adverse reactions. Conversely, a lower value implies that the medication combinations may be safer and reduce unnecessary medication use. Notably, this metric serves as a reference only, and the size of the combination should not be used as a strict evaluation standard.

\subsection{Setup Protocol}
In this subsection, we offer a comprehensive overview of the experimental environment setup, parameter selection, model configuration, and sampling methods for the testing phase.

\subsubsection{Experimental Environment}
The experiments were conducted on an Ubuntu 22.04 system equipped with 30GB of memory, 12 CPUs, and a 24GB NVIDIA RTX3090 GPU, using PyTorch 2.0.0 and CUDA 11.7.

\subsubsection{Configuration and Parameter}
For the embedding matrices of entities \(E_d\), \(E_p\), and \(E_m\), we set \( 128 \) as the embedding size, initialized in the range of -0.1 to 0.1. For the molecular GNN, we implement a 4-layer GIN with a hidden size of 128. For each molecular graph, the initial set of 9-dimensional node features comprise atom count, chirality, and other additional atomic features. The 3-dimensional edge features include bond type, bond stereochemistry, and conjugated bonds. The molecular 3D features include molecular conformations, 3D coordinates, RBF distances, and normalized interatomic vectors. Additionally, we implement a 3-layer GVP, where the hidden dimensions for node feature scalars and vectors are 128 and 64, respectively, and the hidden dimensions for edge feature scalars and vectors are 32 and 1, respectively. For the GRU, we configure the same number of hidden units as the number of medications. For the loss functions, we define the thresholds \(\delta = 0.5\), \(\tau = 2.5\), \(\beta = 0.95\), and \(\phi = 0.06\). We conduct training for 300 epochs during the pretraining stage and 20 epochs during the training stage.

\subsubsection{Sampling Approach}
Given the limited availability of publicly accessible EHR data, we utilize bootstrapping techniques in this phase, following the methodology outlined in \citep{safedrug}. As discussed in \citep{testsample1} and \citep{testsample2}, this technique is particularly effective for small sample size scenarios.

\subsection{Performance Comparison}

\begin{table*}
    \caption{The performance of each model on the test set regarding accuracy and safety. The best and the runner-up results are highlighted in bold and underlined respectively.}
    \begin{tabular}{|*{1}{>{\centering\arraybackslash}p{1.4cm}}| *{4}{>{\centering\arraybackslash}p{1.1cm}} *{1}{>{\centering\arraybackslash}p{1.25cm}} | *{4}{>{\centering\arraybackslash}p{1.1cm}} *{1}{>{\centering\arraybackslash}p{1.25cm}} |}
    \toprule
    \multirow{2}{*}{Model}
    & \multicolumn{5}{c|}{MIMIC-III} & \multicolumn{5}{c|}{MIMIC-IV} \\ 
    \cmidrule(lr){2-6} \cmidrule(lr){7-11}
    & Jaccard$\uparrow$ & DDI$\downarrow$ & F1$\uparrow$ & PRAUC$\uparrow$ & Avg.\#Med & Jaccard$\uparrow$ & DDI$\downarrow$ & F1$\uparrow$ & PRAUC$\uparrow$ & Avg.\#Med \\
    \midrule
    LR          & 0.4924    & 0.0830    & 0.6490    & 0.7548    & 16.0489   & 0.4569    & 0.0783    & 0.6064    & 0.6613    & 8.5746 \\
    ECC         & 0.4856    & 0.0817    & 0.6438    & 0.7590    & 16.2578   & 0.4327    & 0.0764    & 0.6129    & 0.6530    & 8.7934 \\
    RETAIN      & 0.4871    & 0.0879    & 0.6473    & 0.7600    & 19.4222   & 0.4234    & 0.0936    & 0.5785    & 0.6801    & 10.9576 \\ 
    LEAP        & 0.4526    & 0.0762    & 0.6147    & 0.6555    & 18.6240   & 0.4254    & 0.0688    & 0.5794    & 0.6059    & 11.3606 \\
    GAMENet     & 0.4994    & 0.0890    & 0.6560    & 0.7656    & 27.7703   & 0.4565    & 0.0898	& 0.6103	& 0.6829    & 18.5895 \\
    SafeDrug    & 0.5154    & \textbf{0.0655}       & 0.6722    & 0.7627    & 19.4111   & 0.4487	& \textbf{0.0604}	& 0.6014	& 0.6948 & 13.6943 \\
    MICRON      & 0.5219    & 0.0727    & 0.6761    & 0.7489    & 19.2505    & 0.4640    & 0.0691    & 0.6167    & 0.6919    & 12.7701  \\
    COGNet      & \underline{0.5312}    & 0.0839    & 0.6744    & 0.7708    & 27.6335    & 0.4775	& 0.0911	& 0.6233	& 0.6524 & 18.7235 \\
    MoleRec     & 0.5293    & 0.0726    & 0.6834    & 0.7746    & 22.0125      & 0.4744	& 0.0722	& 0.6262	& 0.7124 & 13.4806  \\
    CausalMed     & 0.5389    & 0.0709    & \underline{0.6916}    & \underline{0.7826}    & 20.5419      & \underline{0.4899}	& \underline{0.0677}	& \underline{0.6412}	& \underline{0.7338} & 14.4295  \\
    \midrule
    \textbf{BiMoMed}   & \textbf{0.5422}   & \underline{0.0707}  & \textbf{0.6944}   & \textbf{0.7858}    & 20.3989        & \textbf{0.4917}	 & 0.0696	 & \textbf{0.6425}	   & \textbf{0.7366} & 13.5291\\
    \bottomrule
    \end{tabular}
    \label{tab:comparison}
\end{table*}


In this section, we compare our model with baseline models, assessing model accuracy and safety using key metrics. For baseline methods with a provided test model, we utilize the baseline model directly. For methods without a provided test model, we retrain the models and perform evaluations. We use the optimal parameter settings described in the respective papers to derive results from each baseline method. Table \ref{tab:comparison} presents the comparative results.

\begin{figure}
    \centering
    \includegraphics[width=0.9\linewidth]{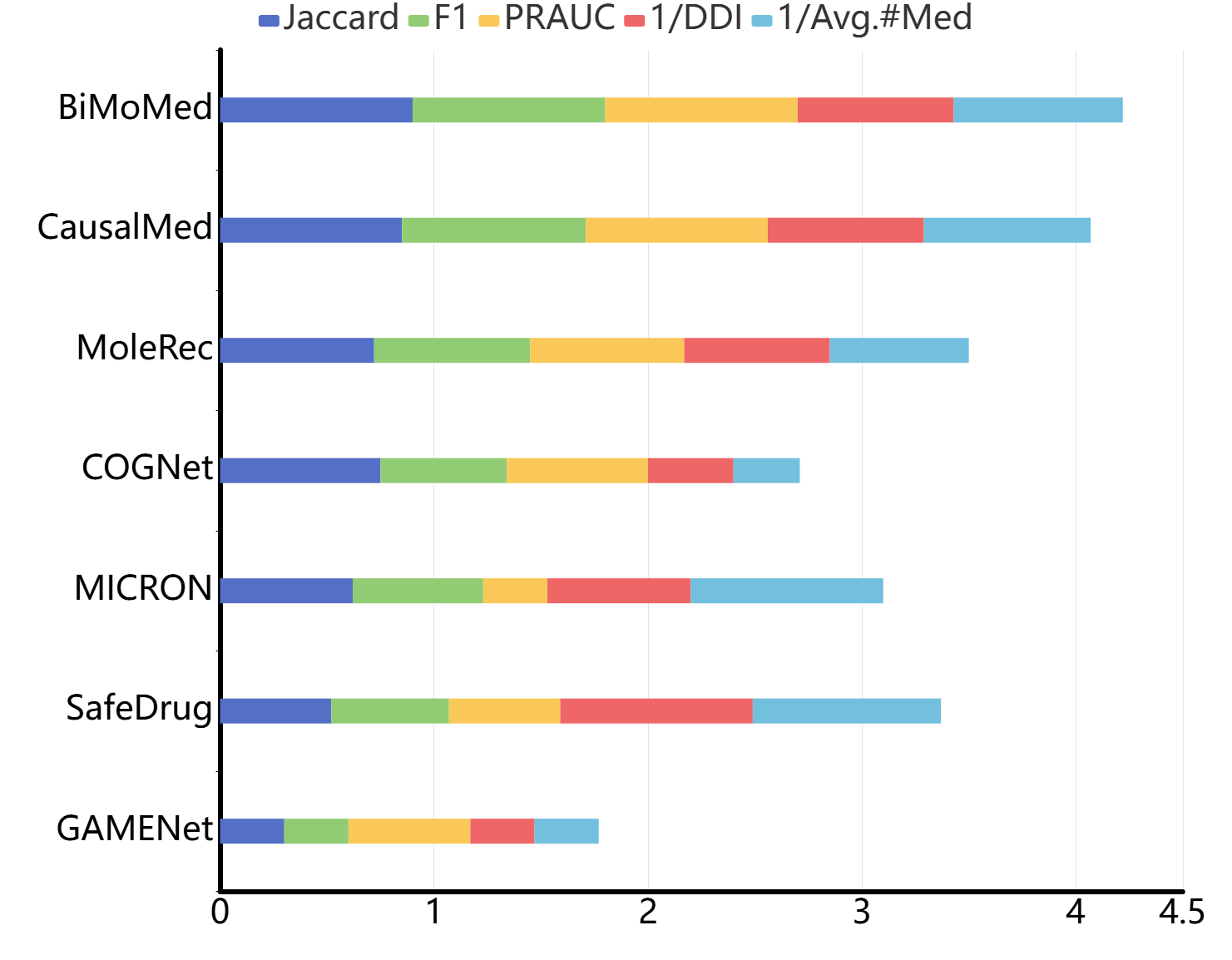}
    \caption{Comparison with recent outstanding works across all metrics in MIMIC-III.}
    \label{fig:comparison}
\end{figure}

To provide a clearer comparison between the baselines and our model, we integrate all evaluation metrics into a single chart. Jaccard, F1, and PRAUC assess model accuracy, while DDI and Avg.\#Med evaluate safety. We normalize the metrics to the 30\%-90\% range and present them in Figure~\ref{fig:comparison}. Notably, the chart does not include the models' efficiency metrics.

\begin{table}
  \caption{The performance of recent excellent models in training and inference efficiency.}
  \begin{tabular*}{\tblwidth}{@{} LCCCC@{} }
   \toprule
    Method & \makecell{Convergence\\ Epoch}  & \makecell{Training Time\\/Epoch(s)} & \makecell{Total Training\\Time(s)} & \makecell{Inference\\ Time(s)} \\
   \midrule
    GAMENet & 39 & 45.31 & 1767.09 & 19.27\\
    SafeDrug & 54 & 38.32 & 2069.28 & 20.15\\
    MICRON & 40 & 17.48 & 699.20 & 14.48\\ 
    COGNet & 103 & 38.85 & 4001.55 & 142.91\\
    MoleRec & 25 & 249.32 &	6233.00 & 32.10\\
    CausalMed & 33 & 164.77 & 5437.41 & 18.29  \\
    \midrule
    BiMoRec & 15 & 416.61 & 6249.15 & 15.16\\
   \bottomrule
  \end{tabular*}
  \label{tab:time}
\end{table}

We also evaluate the operational efficiency of the models, including the number of convergence epochs, training time per epoch, total training time, and inference time, as outlined in Table \ref{tab:time}.

Among the various baseline methods, LR and ECC, which employ traditional recommendation techniques, demonstrate low accuracy. LEAP decomposes treatment recommendations into a sequential decision process, leveraging sequential relationships to enhance effectiveness, but it does not surpass traditional methods. RETAIN also introduces a sequence model but leads to an excessively high DDI rate.

GAMENet significantly improves model accuracy compared to previous models by incorporating patient visit history into the model framework without notably increasing complexity. However, it suffers from a high DDI rate.
SafeDrug significantly reduces the DDI rate by explicitly modeling relationships between medication molecules, but it does not capture relationships between medications and diseases, thereby leaving accuracy improvement potential.
MICRON adjusts recommended medication combinations by capturing changes in patients' physical conditions between two visits, resulting in improved accuracy and training efficiency after certain data filtering.
COGNet designs a replication mechanism that significantly improves the Jaccard metric while increasing the number of iterations, thereby affecting time efficiency and compromising the safety of some medication combinations.
MoleRec incorporates external molecular knowledge by establishing relationships between medication molecules, their substructures, and patients' diseases and procedures. This approach enhances accuracy without significantly compromising safety. However, it does not account for interactions between molecules or the effects of single molecular modalities. 
CausalMed recommends direct causal relationships by exploring the causal connections between medical entities, which not only improves accuracy but also enhances the interpretability of medication recommendations. Nonetheless, it does not consider relationships between entities at the molecular level.

Our model introduces the 3D modality of molecules, recalibrating the representations of molecules and their substructures, thereby enhancing molecular dimensionality and resolving structural ambiguity. We design a molecular multi-step enhancement mechanism, refining the interaction between molecular knowledge and the patient's health condition, improving the model's logic and accuracy. This approach only slightly increases the DDI rate, maintaining a balance between accuracy and safety.

In terms of operational efficiency, although the training time per epoch for our model is relatively long, the number of epochs required for convergence is greatly reduced in practice, often needing only 15 epochs. As a result, the total training time remains comparable to MoleRec \citep{molerec}, which also utilizes molecular knowledge.

\section{Discussions}
In this section, we conduct an in-depth analysis of the experimental outcomes previously discussed. Additionally, we carry out a series of supplementary experiments to confirm the comprehensiveness and rationale of our approach, to ensure its robustness and validate its advancements in the field.

\subsection{Effectiveness Analysis}
The comparative experiments above showcase the superior performance of our model in terms of both accuracy and safety.

Previous studies on medication recommendation typically base their recommendations on the patient’s physical condition. With the introduction of external molecular knowledge, both accuracy and safety are further enhanced. However, the molecular structures introduced are merely 2D projections of the actual molecular configurations, resulting in structural ambiguity. Furthermore, molecular substructures fail to independently interact with the state of a single patient visit, leading to the absence of crucial substructure combinations necessary for each visit's medication.

To address these issues, we introduce the 3D modality of molecules to supplement the missing three-dimensional information in molecular representations, thereby resolving the issue of structural ambiguity. Simultaneously, we design a molecular multi-step enhancement mechanism to optimize the use of molecular embeddings, capturing the interactions within the molecular set during a single visit and making critical molecular adjustments for each patient visit. Finally, we implement the final enhancement in response to the patient's most recent condition. Our proposed method addresses issues overlooked by other models, substantially improving the overall efficacy of the medication recommendation system.

\subsection{Efficiency Analysis}
To more accurately represent the efficiency of our model, we conduct a stepwise analysis of the model's time complexity. In the pre-training module, we use GIN and GVP networks to generate representations of two molecular modalities. The time complexity of the GNN layer is \(O((G_{n}+G_{e})^4)\), with \(G_n\) representing the number of nodes, \(G_e\) the number of edges, and 4 denoting the number of layers. The time complexity of the NTXent layer is \(O(n^2)\), with \(n\) representing the number of nodes. Thus, the total time complexity of the pre-training phase is approximately \(O((G_{n}+G_{e})^4 + n^2)\), where \(O(n^2)\) can be ignored.

In the subsequent multi-step molecular enhancement phase, embedding layers, GRU layers, molecular structure and substructure graph neural networks, attention mechanisms, and multiple fully connected layers are employed for computation. The time complexity of the embedding layer is \(O(L \cdot dim)\), with \(L\) representing the input sequence length and \(\text{dim}\) the size of the embedding dimension. The time complexity of the GRU layer is \(O(g \cdot h^2)\), where \(g\) denotes the sequence length and \(h\) the hidden layer size. The GNN layer exhibits a time complexity of \(O((G_{n}+G_{e})^4)\), and the attention mechanism has a time complexity of \(O(L^2 \cdot dim)\). The time complexity of the fully connected layers is represented as \(O(\Sigma (n_i \cdot n_o))\), with \(n_i\) representing the input nodes and \(n_o\) the output nodes. Thus, the overall time complexity is:
\begin{equation}
    O((L + L^2) \cdot dim + g \cdot h^2 + (G_{n}+G_{e})^4 + \Sigma(n_i \cdot n_o)).
\end{equation}

Through these calculations, we observe that the number of layers in the GNN network is a critical factor in model configuration. Although a 4-layer network structure increases model runtime, it facilitates faster model convergence. Thus, overall time efficiency remains unaffected.

\subsection{Ablation Study}
To evaluate the contribution of our innovations to the overall performance of the model and minimize interference from external factors, we remove key modules and make logical modifications to set up several variant experiments.

\begin{table*}
\caption{The performance of each ablation model on the test set regarding accuracy and safety. The best and the runner-up results are highlighted in bold and underlined respectively.}
    \begin{tabular}{|*{1}{>{\centering\arraybackslash}p{2.7cm}}| 
    *{4}{>{\centering\arraybackslash}p{0.95cm}} 
    *{1}{>{\centering\arraybackslash}p{1.15cm}} |
    *{4}{>{\centering\arraybackslash}p{0.95cm}} 
    *{1}{>{\centering\arraybackslash}p{1.15cm}} |}
    \toprule
    \multirow{2}{*}{Model}
    & \multicolumn{5}{c|}{MIMIC-III} & \multicolumn{5}{c|}{MIMIC-IV} \\ 
    \cmidrule(lr){2-6} \cmidrule(lr){7-11}
    & Jaccard$\uparrow$ & DDI$\downarrow$ & F1$\uparrow$ & PRAUC$\uparrow$ & Avg.\#Med & Jaccard$\uparrow$ & DDI$\downarrow$ & F1$\uparrow$ & PRAUC$\uparrow$ & Avg.\#Med \\
    \midrule
    BiMoRec $w/o$ T        & \underline{0.5356}	& \textbf{0.0697}	& \underline{0.6887}	& \underline{0.7823}	& 20.4086   & \underline{0.4832}	& 0.0704	& \underline{0.6348}	& \underline{0.7268}	& 13.5401  \\
    BiMoRec $w/o$ E        & 0.5347	& 0.0735	& 0.6882	& 0.7752	& 21.8691    & 0.4800    & 0.0698    & 0.6320	& 0.7231	& 15.5599 \\
    BiMoRec $w/o$ T+E      & 0.5335	& 0.0747    & 0.6873	& 0.7782	& 22.7640   & 0.4778	& \underline{0.0696}	& 0.6301	& 0.7212	& 15.9020 \\
    \midrule
    \textbf{BiMoRec}   & \textbf{0.5422}   & \underline{0.0707}  & \textbf{0.6944}   & \textbf{0.7858}    & 20.3989        & \textbf{0.4917}	 & \textbf{0.0696}	 & \textbf{0.6425}	   & \textbf{0.7366} & 13.5291\\
    \bottomrule
    \end{tabular}
    \label{tab:ablation}
\end{table*}


\textbf{BiMoRec $w/o$ T}: In this variant, we remove the pre-training module, thereby omitting the use of the model's 3D modality data. As with prior approaches, we use the SMILES string of molecules to construct graph data as input for the graph neural network, to generate embeddings of the molecular structure and its substructures.

\textbf{BiMoRec $w/o$ E}: In this variant, we removed the molecular multi-step enhancement module, thereby eliminating the interaction between molecules and single patient visits, along with intermolecular and latest visit enhancements. Instead, we used aggregated patient representations derived from comprehensive patient histories to identify the required molecules, which were then transformed into probabilities for generating medication recommendations.

\textbf{BiMoRec $w/o$ T+E}: In this variant, we concurrently remove both the pre-training module and the molecular multi-step enhancement module. This variant does not integrate the dual modalities of molecular representation, nor does it interact with the patient's single representation, relying solely on the relationships between medications, diseases, and procedures in patient visits to make recommendations.

As shown in Table \ref{tab:ablation}, the ablation experiment results align with our expectations.

BiMoRec $w/o$ T demonstrates that by introducing the 3D molecular modality, it improves recommendation accuracy, supplements the missing three-dimensional spatial information of the molecules, resolves structural ambiguity, and optimizes the molecular representation space, as further substantiated in the subsequent case analysis.

BiMoRec $w/o$ E shows that our multi-step enhancement module captures relevant molecular substructures during each visit, thereby enhancing the model’s performance. This demonstrates that capturing the relationship between patient representations and molecular structures at visit-level granularity is advantageous for model performance.

BiMoRec $w/o$ T+E exhibits a slight improvement in safety but a noticeable decrease in accuracy. This phenomenon is common in medication recommendation systems, as there is often a necessary balance between safety and accuracy, with the best possible performance typically resulting from a trade-off between the two. Furthermore, the experiment demonstrates that the molecular bimodal mechanism and the multi-step enhancement mechanism, along with patient representation interactions, exhibit a synergistic effect, and their combined use further enhances model performance.

\subsection{Parameter Sensitivity}

\begin{figure*}
    \centering
    \begin{subfigure}{\linewidth}
        \centering
        \includegraphics[width=0.24\textwidth]{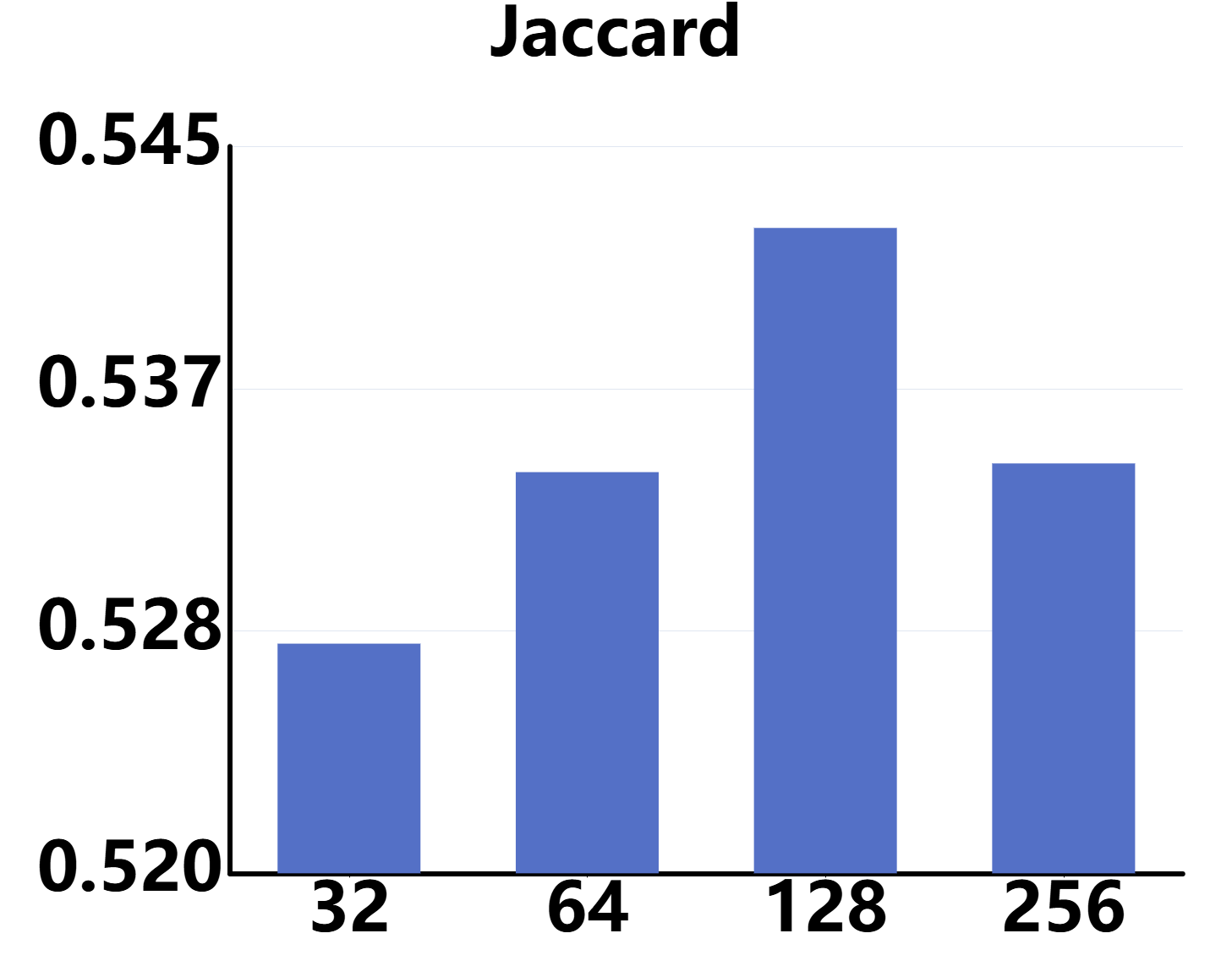}
        \includegraphics[width=0.24\textwidth]{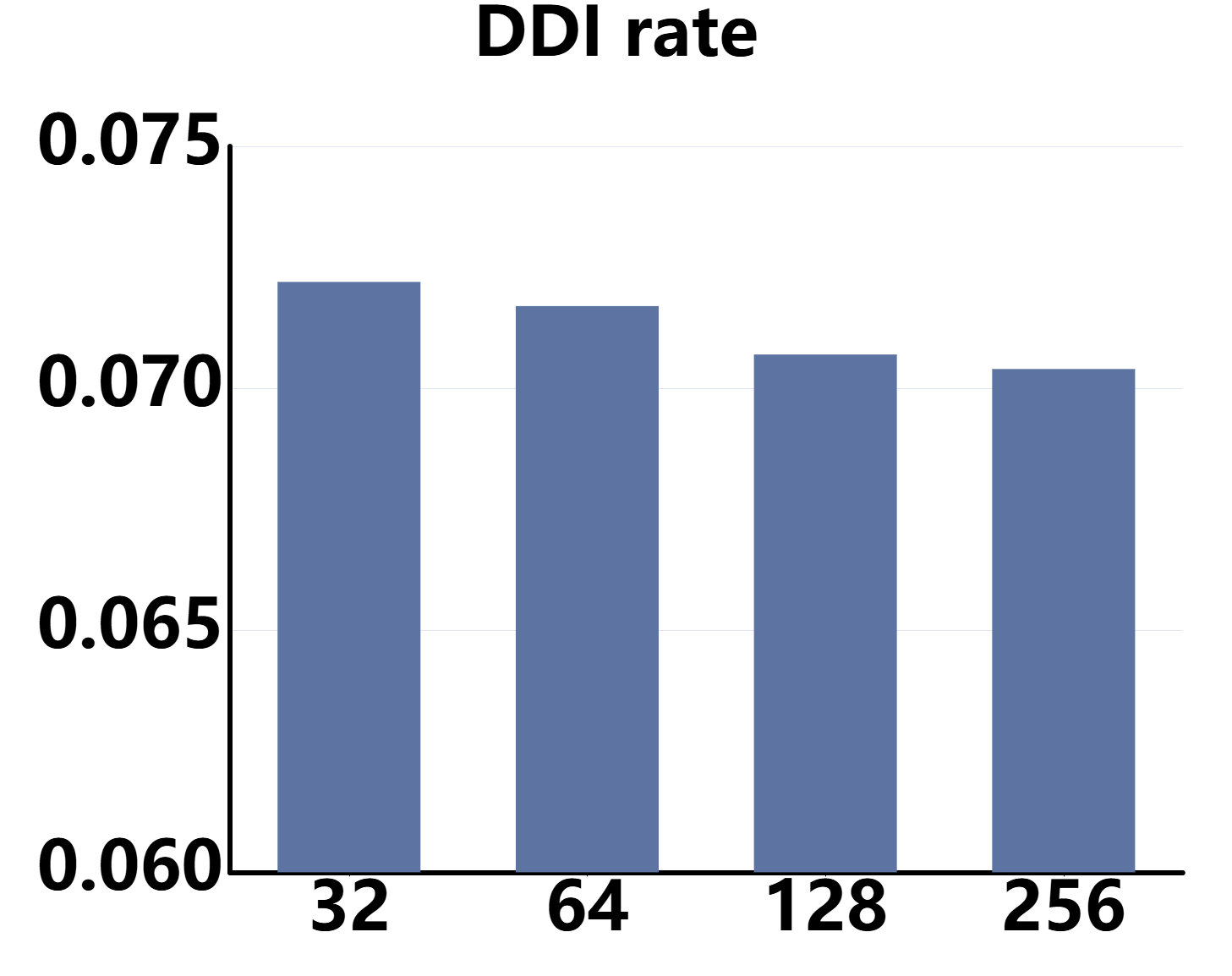}
        \includegraphics[width=0.24\textwidth]{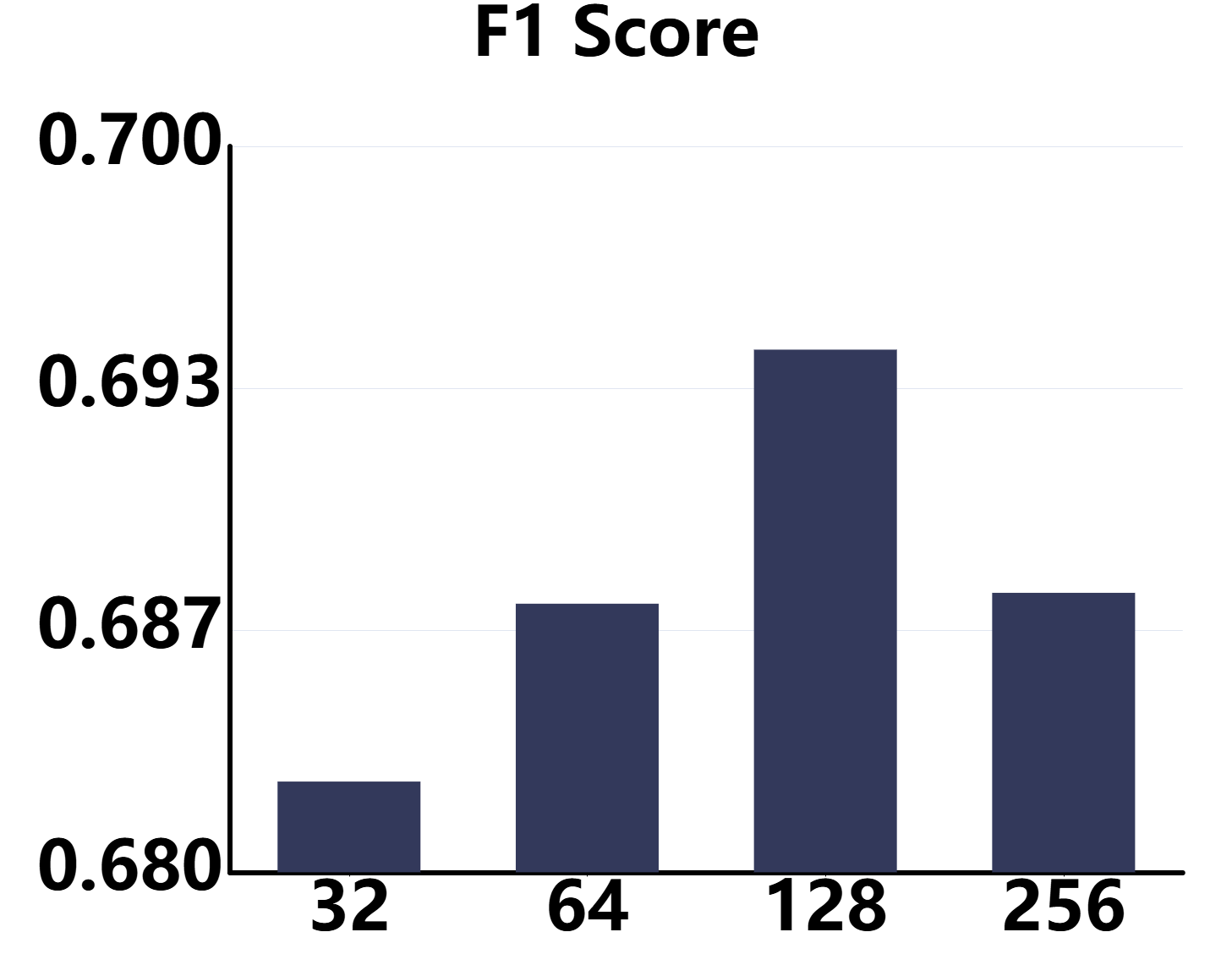}
        \includegraphics[width=0.24\textwidth]{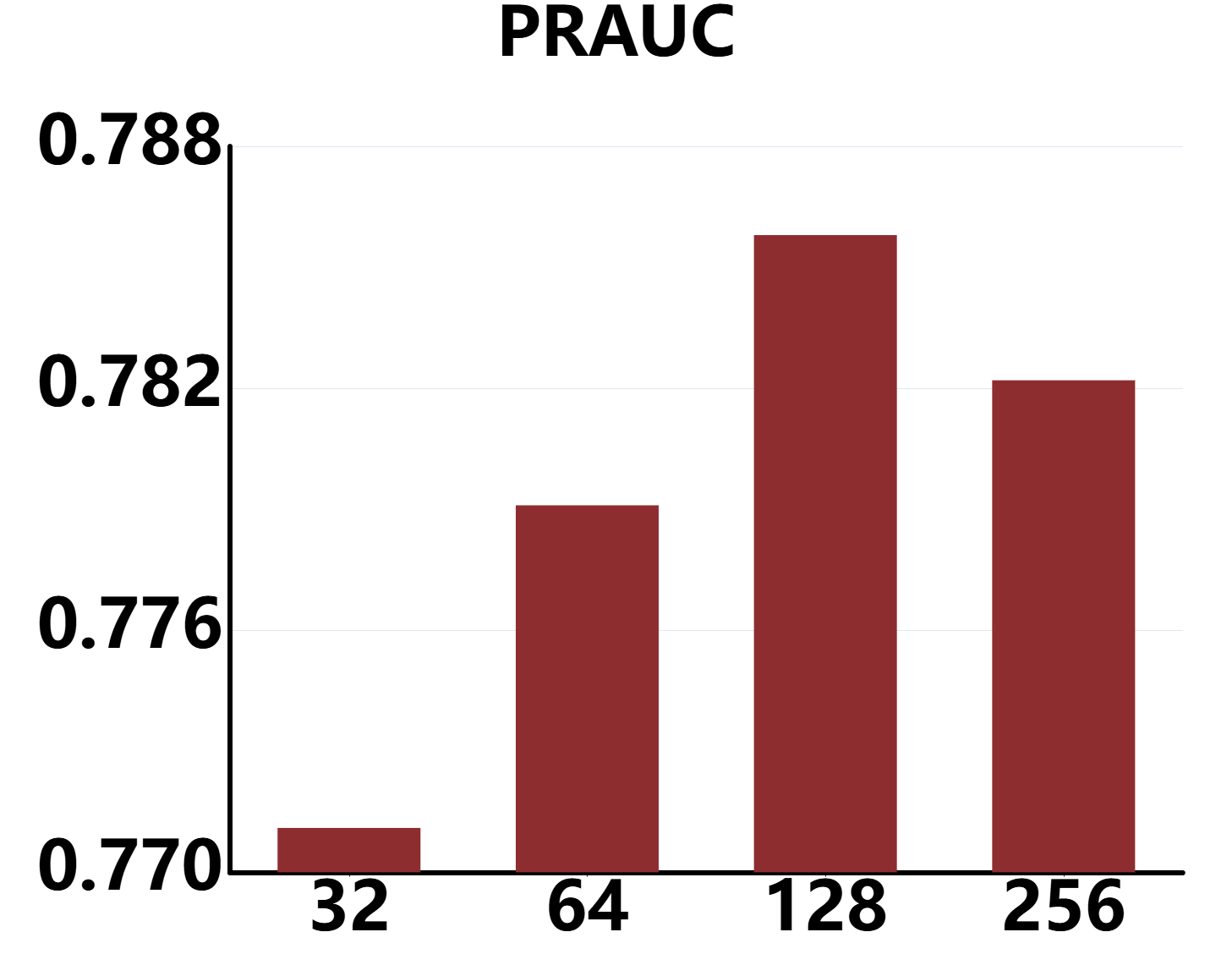}
        \caption{The impact of embedded dimensions on results.}
        \label{fig:para_embedding}
    \end{subfigure}

    \begin{subfigure}{\linewidth}
        \centering
        \includegraphics[width=0.24\textwidth]{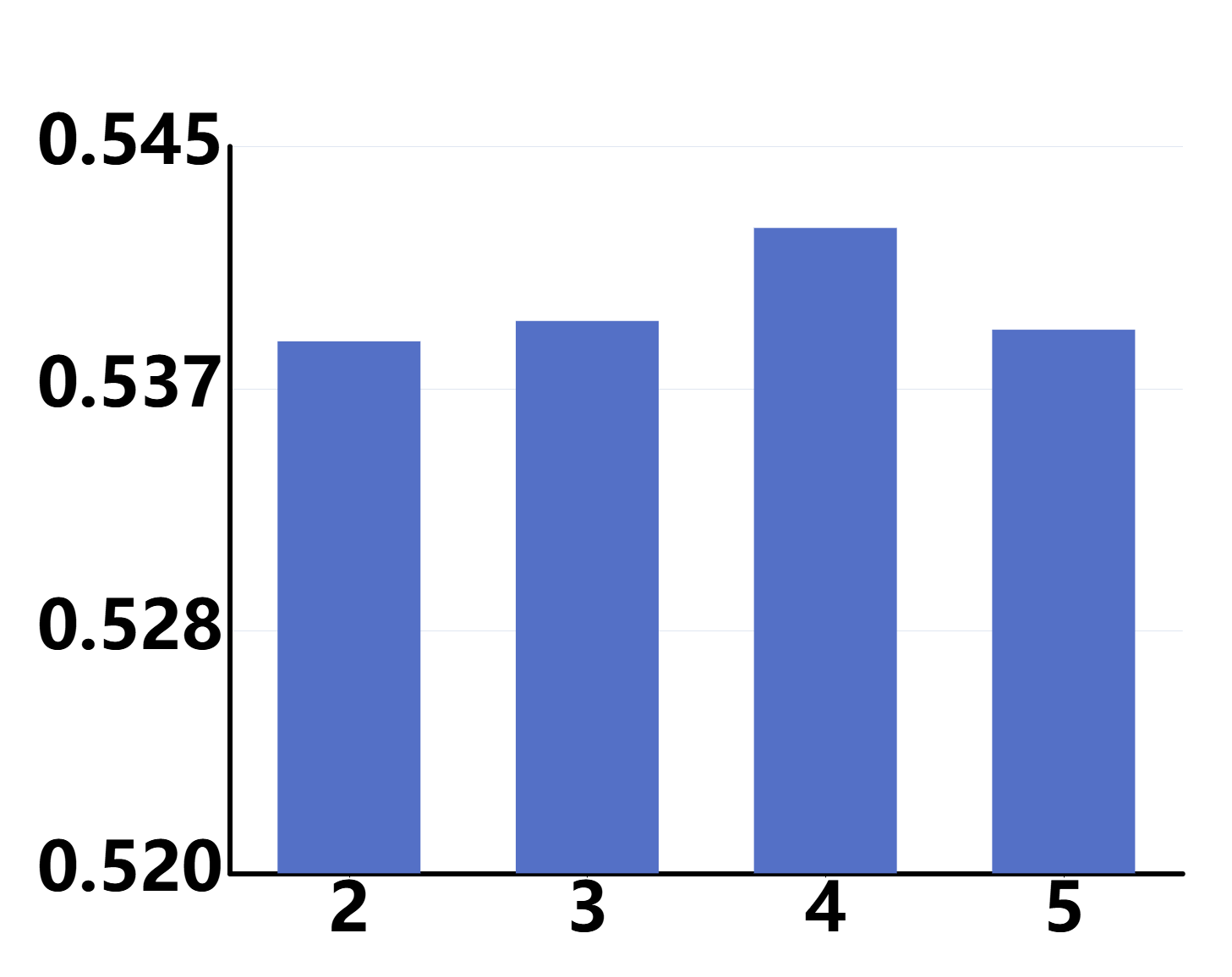}
        \includegraphics[width=0.24\textwidth]{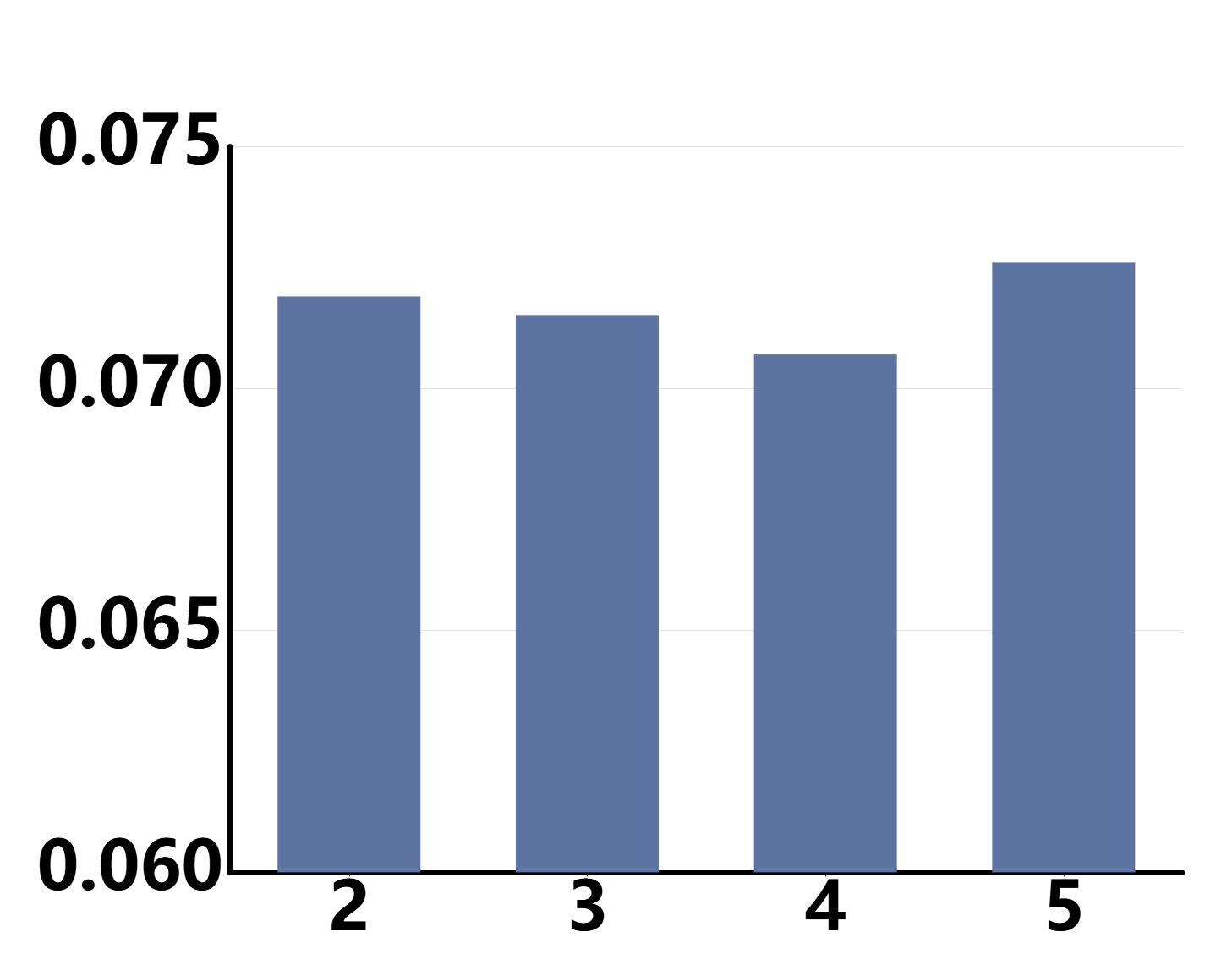}
        \includegraphics[width=0.24\textwidth]{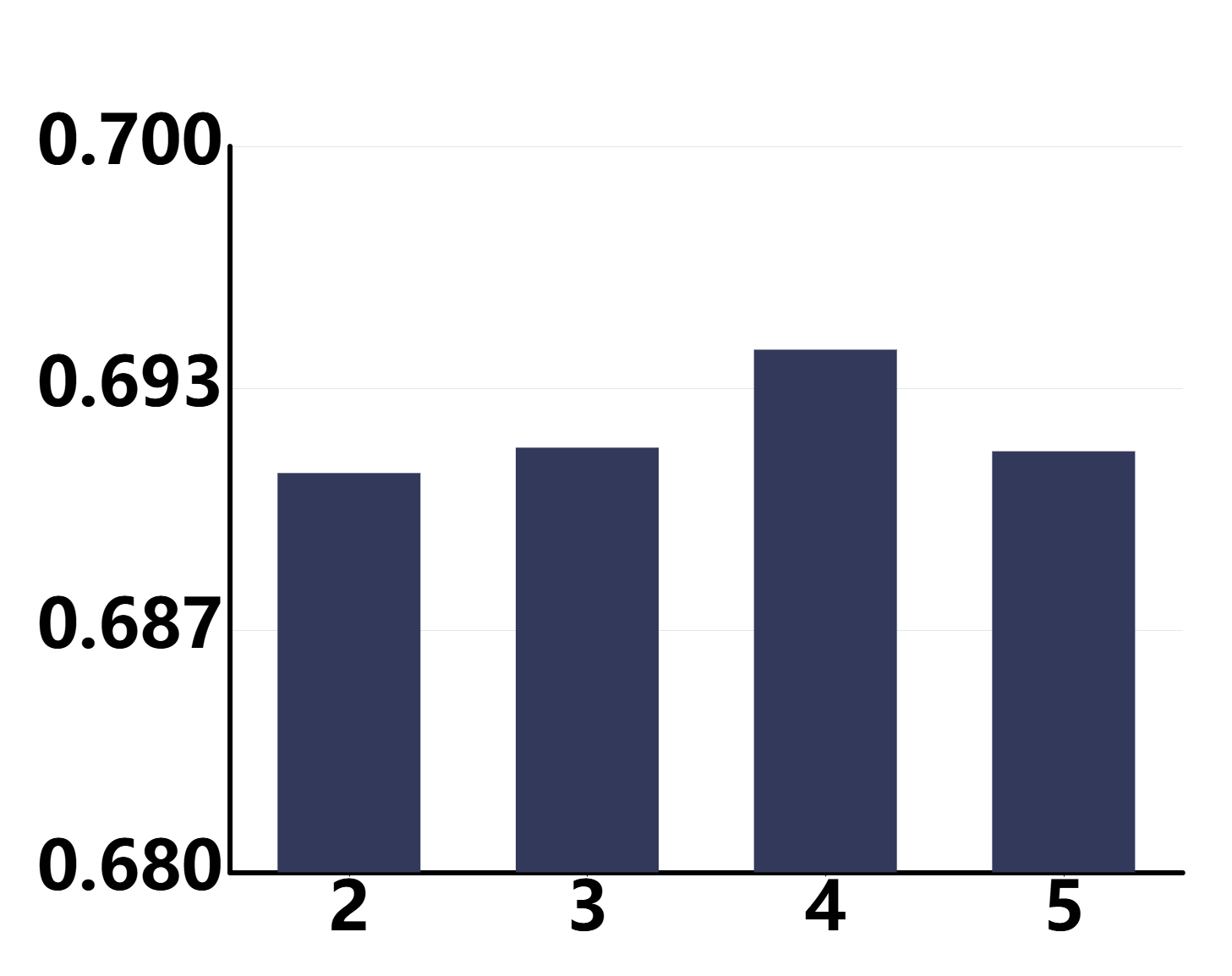}
        \includegraphics[width=0.24\textwidth]{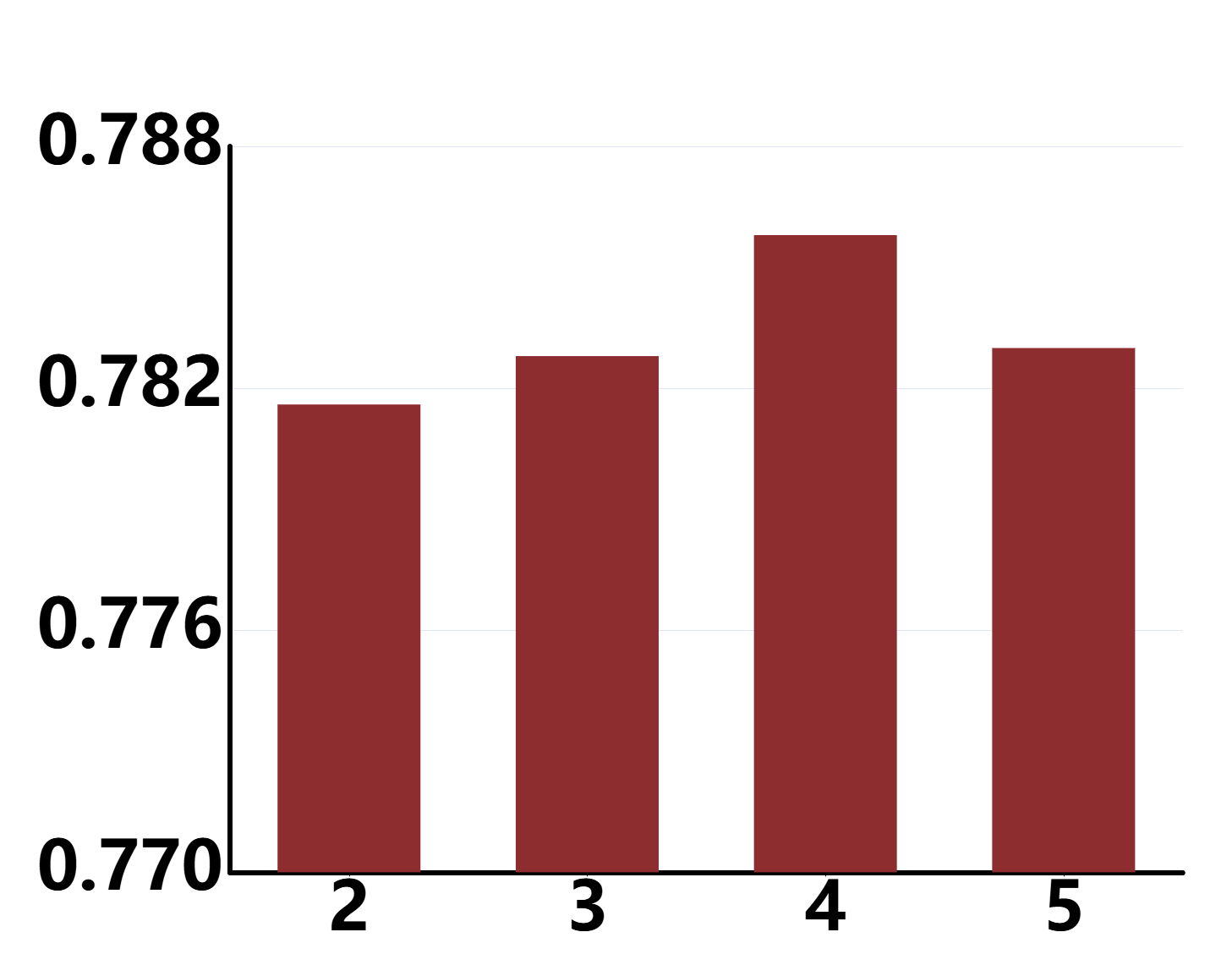}
        \caption{The impact of GNN layers' number on results.}
        \label{fig:para_gnnLayer}
    \end{subfigure}
    
    \begin{subfigure}{\linewidth}
        \centering
        \includegraphics[width=0.24\textwidth]{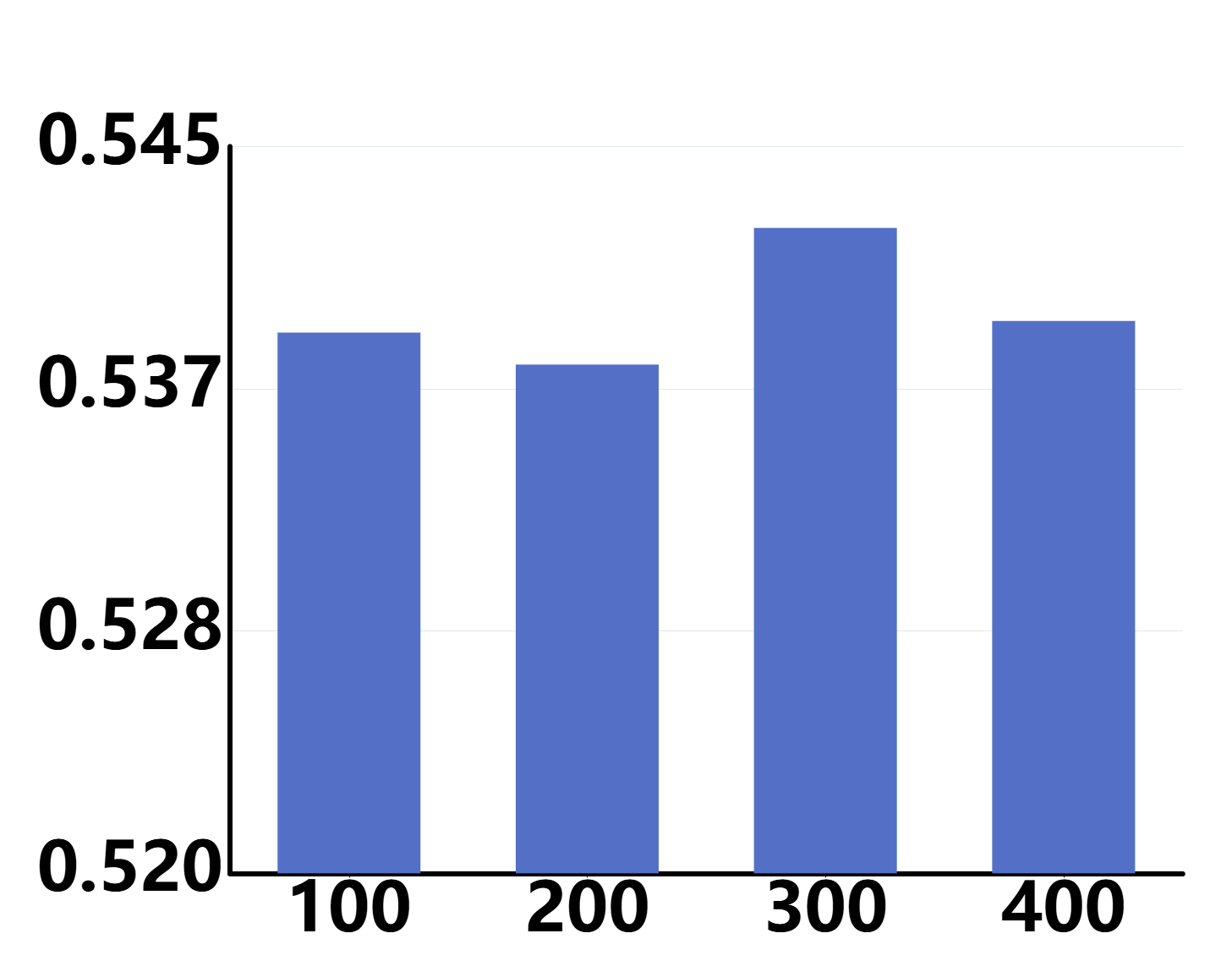}
        \includegraphics[width=0.24\textwidth]{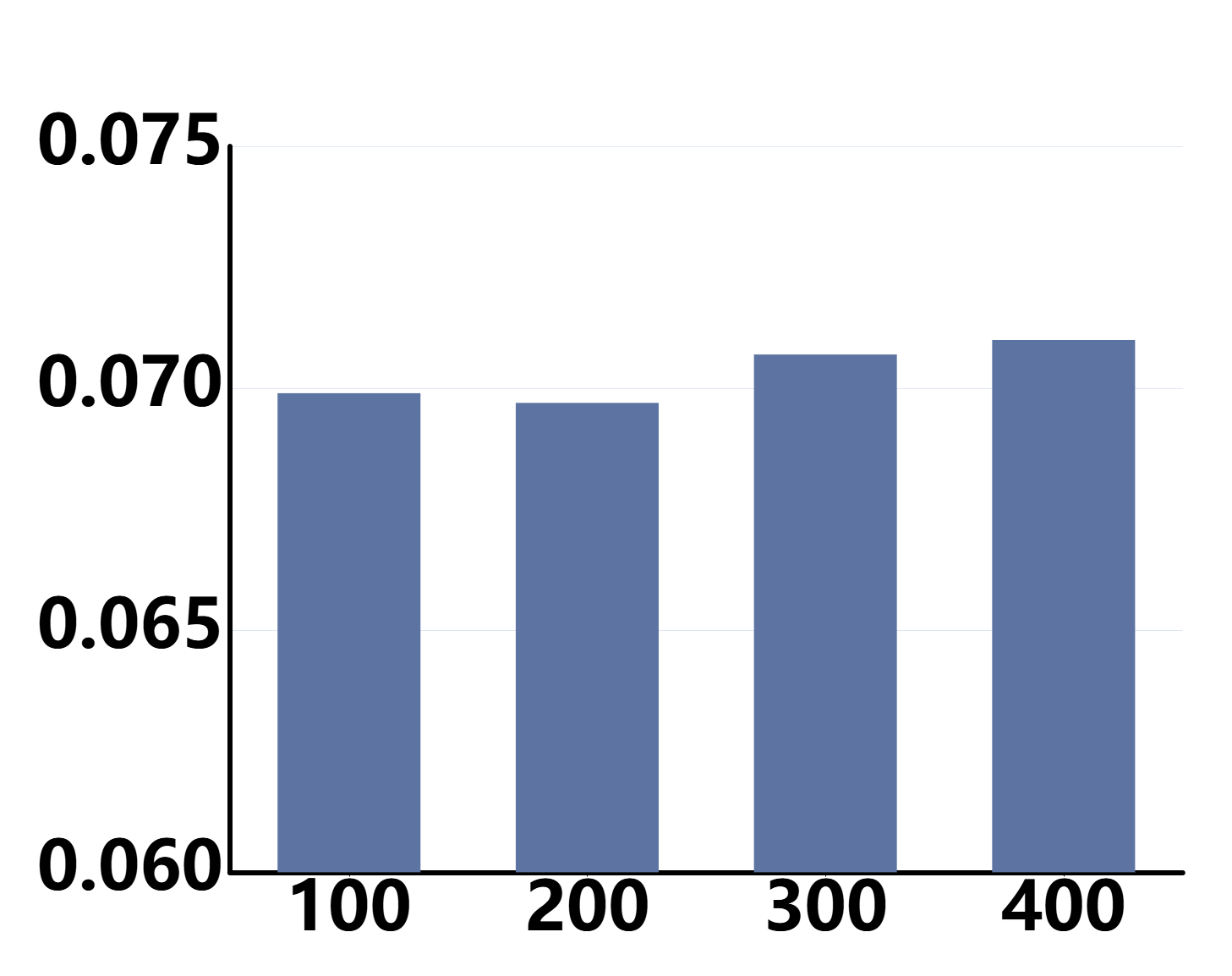}
        \includegraphics[width=0.24\textwidth]{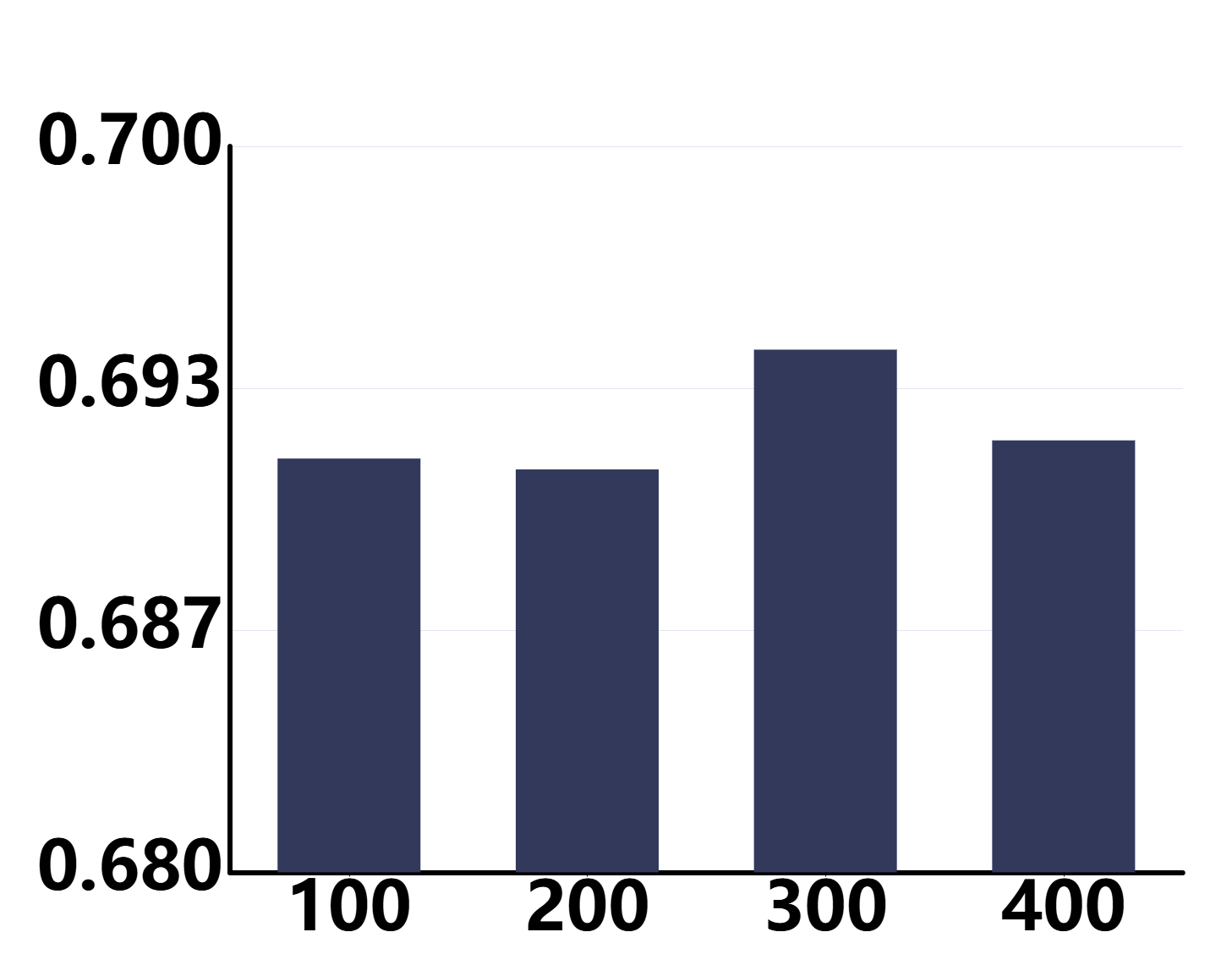}
        \includegraphics[width=0.24\textwidth]{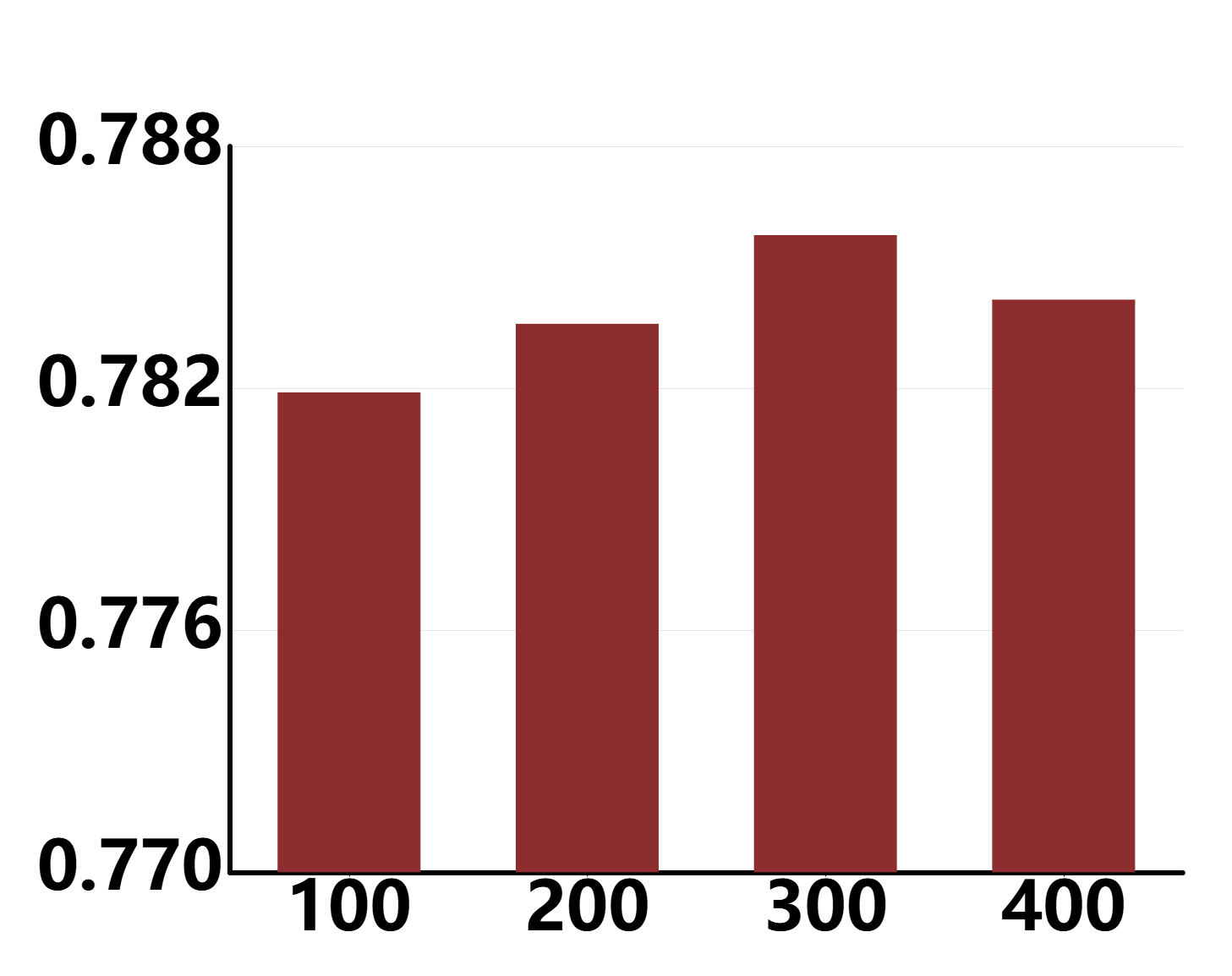}
        \caption{The impact of pre-training epochs on results.}
        \label{fig:para_epoch_pretrain}
    \end{subfigure}

    \begin{subfigure}{\linewidth}
        \centering
        \includegraphics[width=0.24\textwidth]{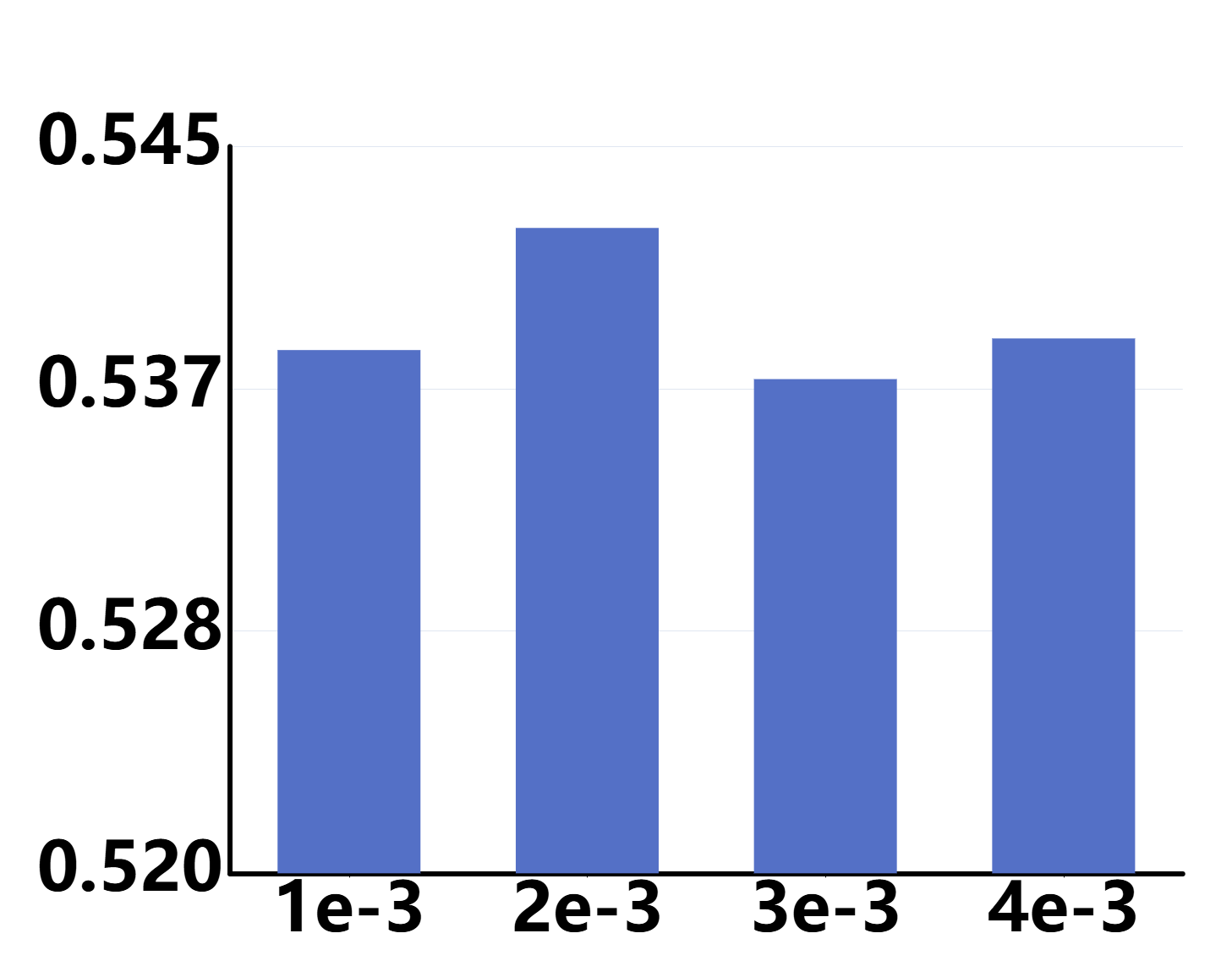}
        \includegraphics[width=0.24\textwidth]{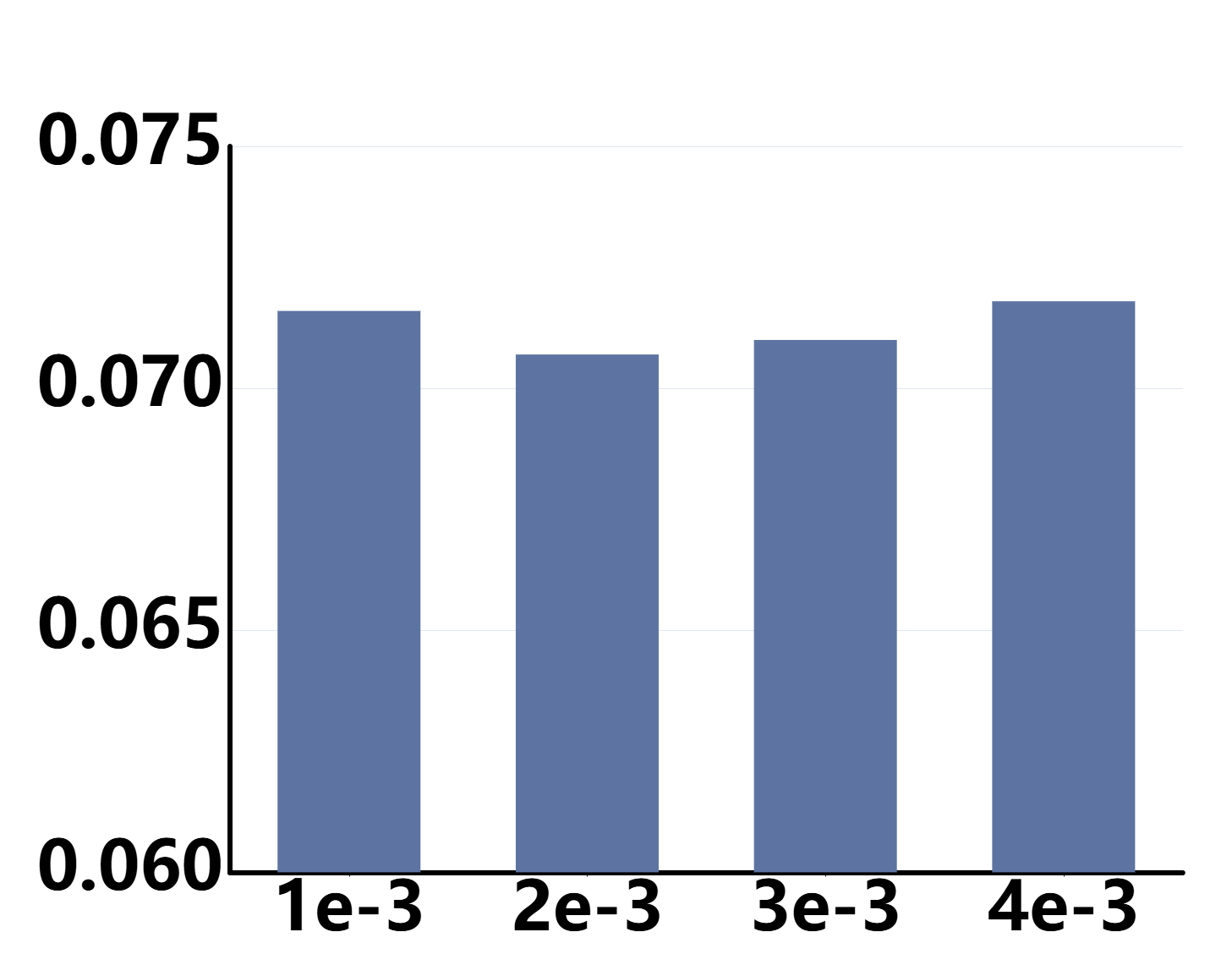}
        \includegraphics[width=0.24\textwidth]{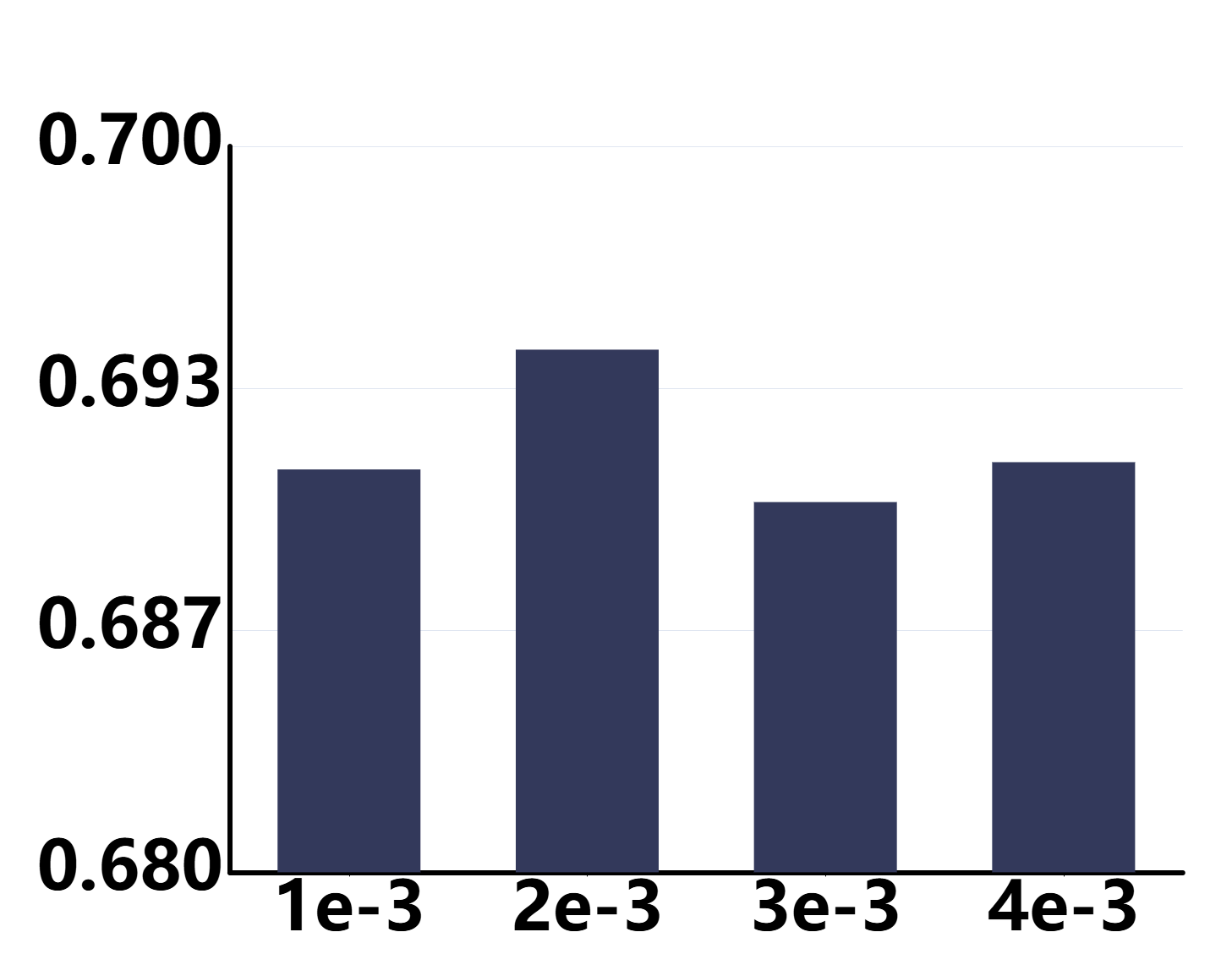}
        \includegraphics[width=0.24\textwidth]{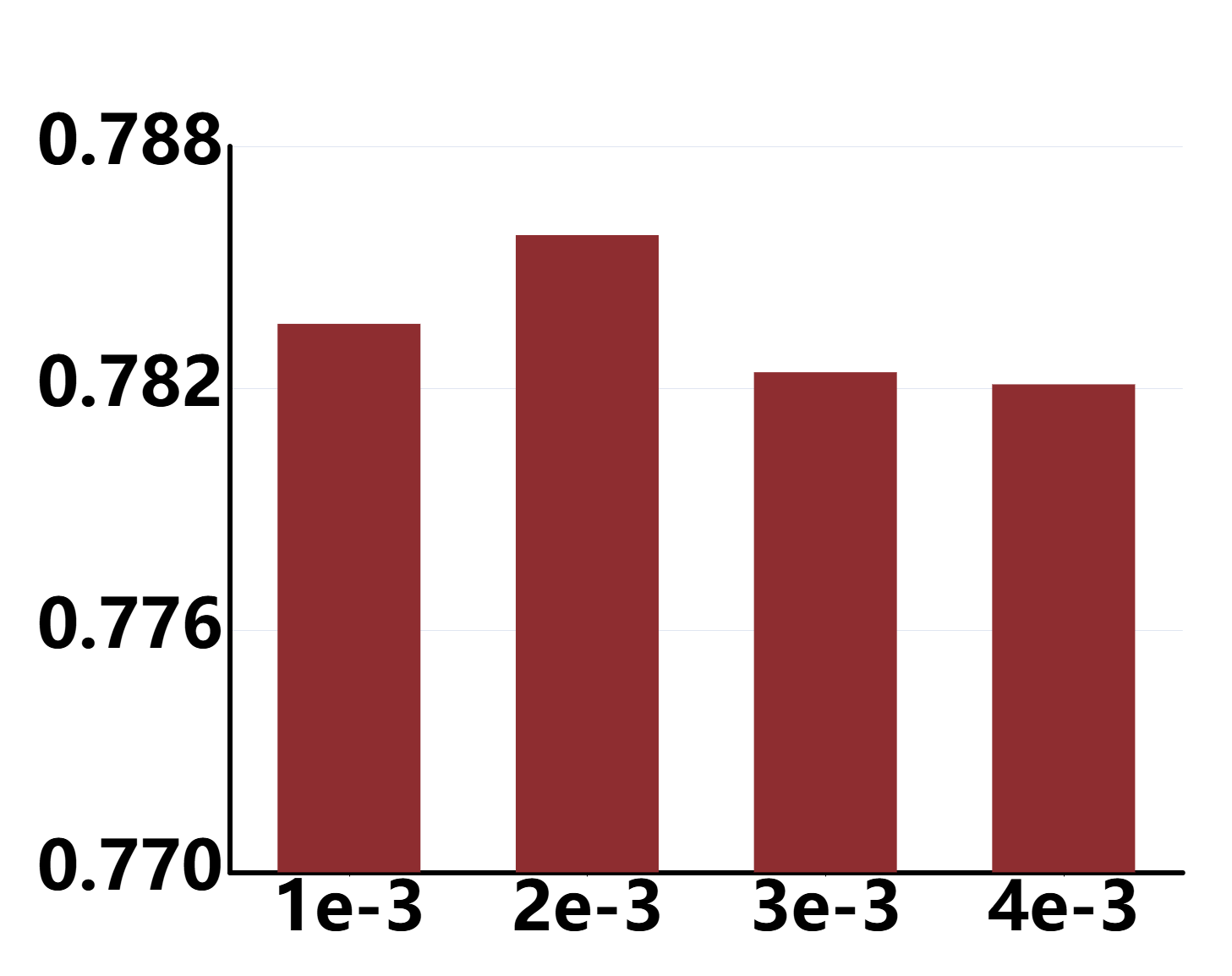}
        \caption{The impact of pre-training learning rate on results.}
        \label{fig:para_lr_pretrain}
    \end{subfigure}


    \vspace{-0.2cm}
    \caption{Experiments on parameter sensitivity on MIMIC-III dataset.}
    \label{fig:para_sensitive}
    \vspace{-0.3in}
\end{figure*}

To investigate the influence of specific parameters on the model, we conduct a series of parameter sensitivity experiments on the MIMIC-III dataset, focusing on four key model parameters. These experiments examine the effects of different parameter combinations on model performance and analyze the underlying causes of these changes in detail. The results are depicted in Figure~\ref{fig:para_sensitive}, where the X-axis represents different parameter values, and the Y-axis shows the performance variations caused by these values.

First, we examine the effect of different embedding dimensions on model performance. Figure~\ref{fig:para_embedding} illustrates performance at embedding dimensions of 32, 64, 128, and 256. When the embedding size is 128, the model achieves significantly higher accuracy, while safety remains consistent. Therefore, considering accuracy, safety, and computational cost, we select 128 as the optimal embedding dimension for this experiment.

Next, since both molecular modalities use GNN layers as the basic encoder, we assess the impact of the number of GNN layers. Figure~\ref{fig:para_gnnLayer} shows that a 4-layer network provides superior accuracy. However, too many layers increase model complexity and the likelihood of overfitting, leading to a larger gap between training and testing results. Conversely, fewer layers reduce convergence speed, preventing the model from achieving optimal performance. Thus, we select 4 layers for the model.

Regarding the number of epochs and learning rate of the pre-training model, these are critical to the performance of the pre-trained model. As shown in Figure~\ref{fig:para_epoch_pretrain} and Figure~\ref{fig:para_lr_pretrain}, too few epochs do not sufficiently perform contrastive learning for the dual molecular modalities, while excessive epochs offer diminishing returns and may reduce performance. The pre-training learning rate also requires fine-tuning, and an optimizer strategy for adjusting the learning step size is recommended.

In practical model experiments, parameter adjustments may be required based on the experimental environment and actual results. Different experimental environments may lead to minor variations.

    \subsection{Case Study}
\begin{figure*}
    \centering
    \includegraphics[width=0.49\textwidth]{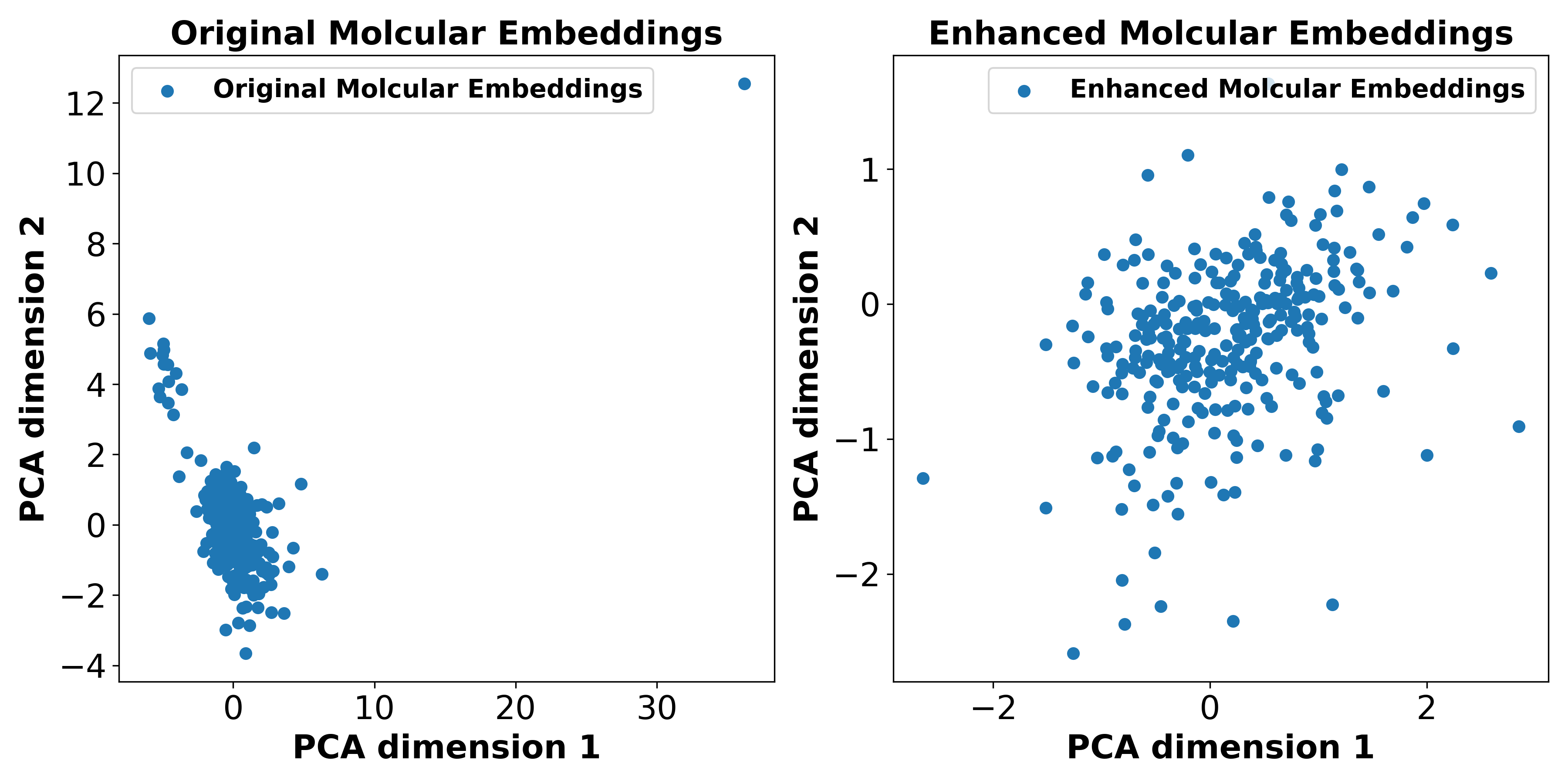}
    \includegraphics[width=0.49\textwidth]{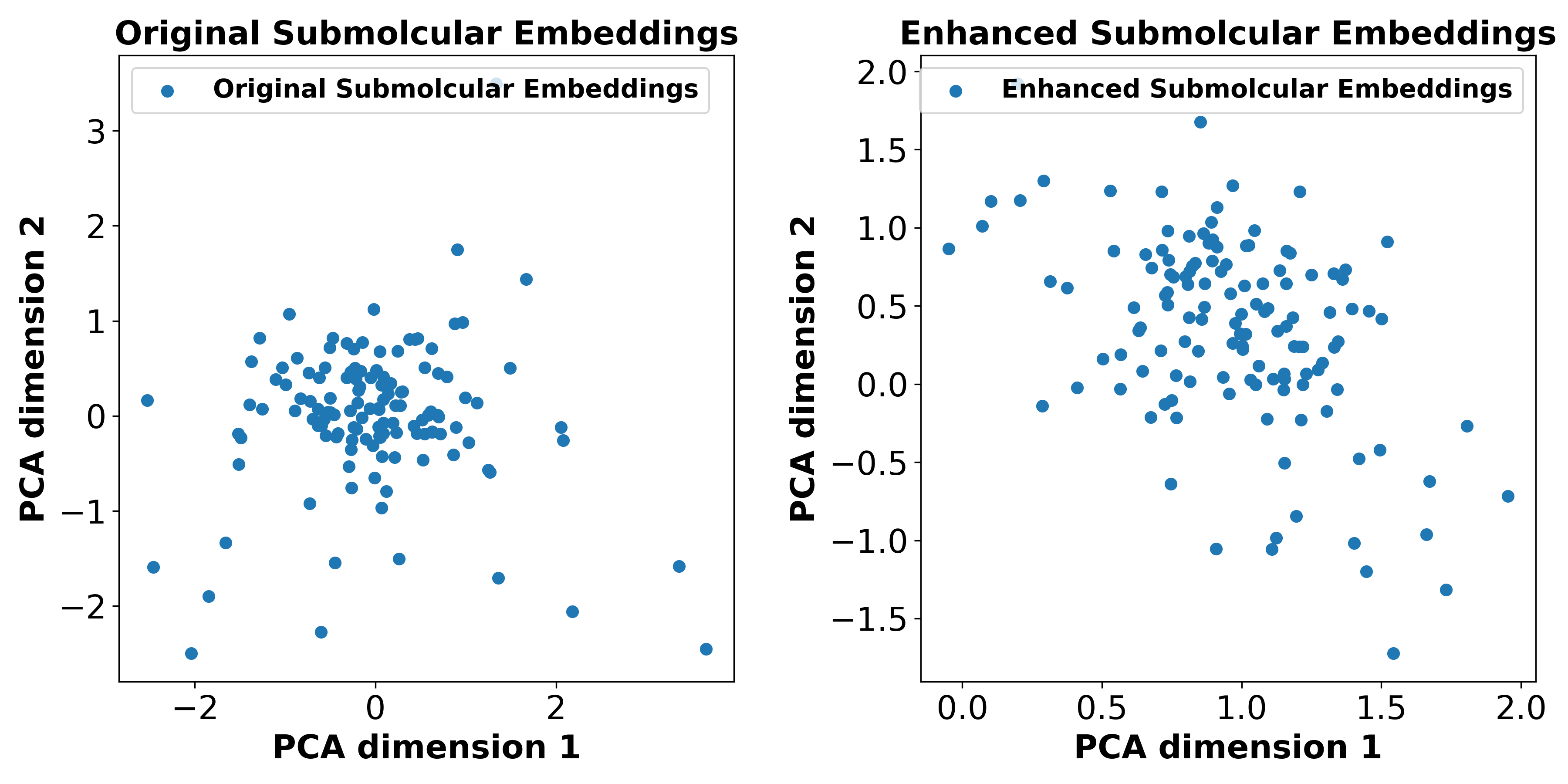}
    \includegraphics[width=0.49\textwidth]{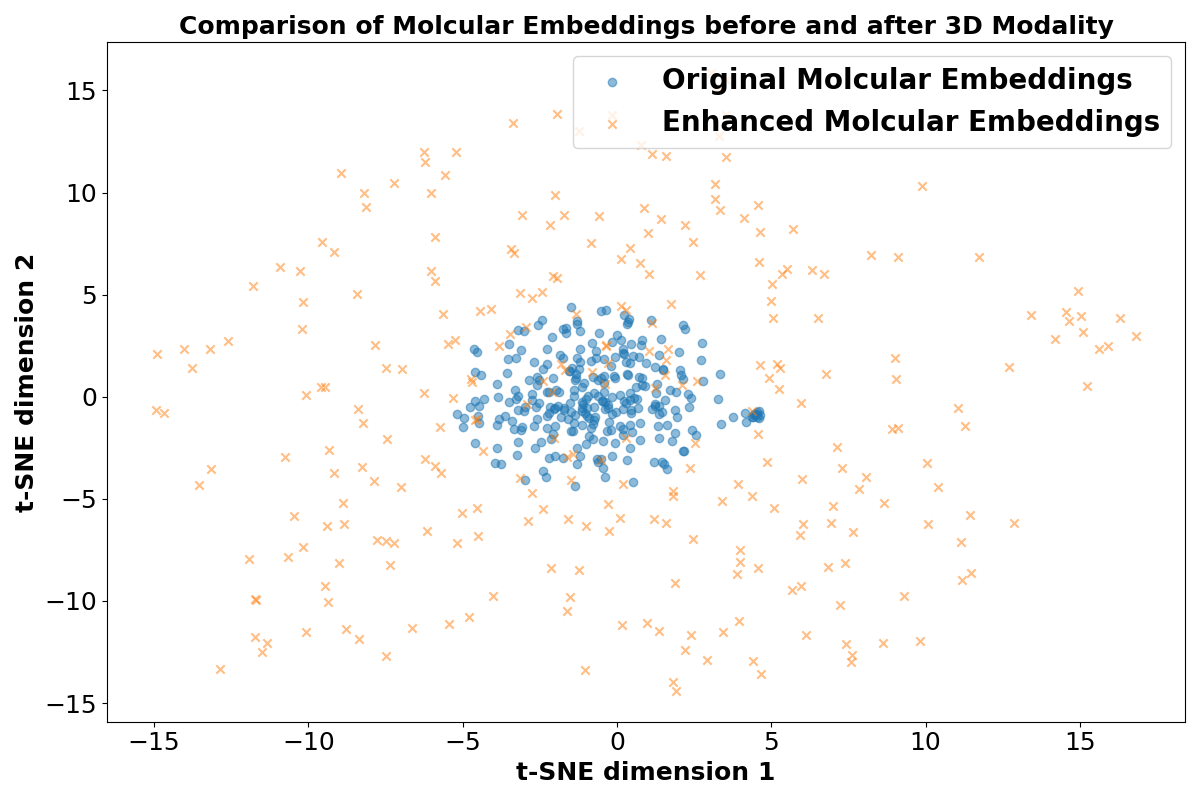}
    \includegraphics[width=0.49\textwidth]{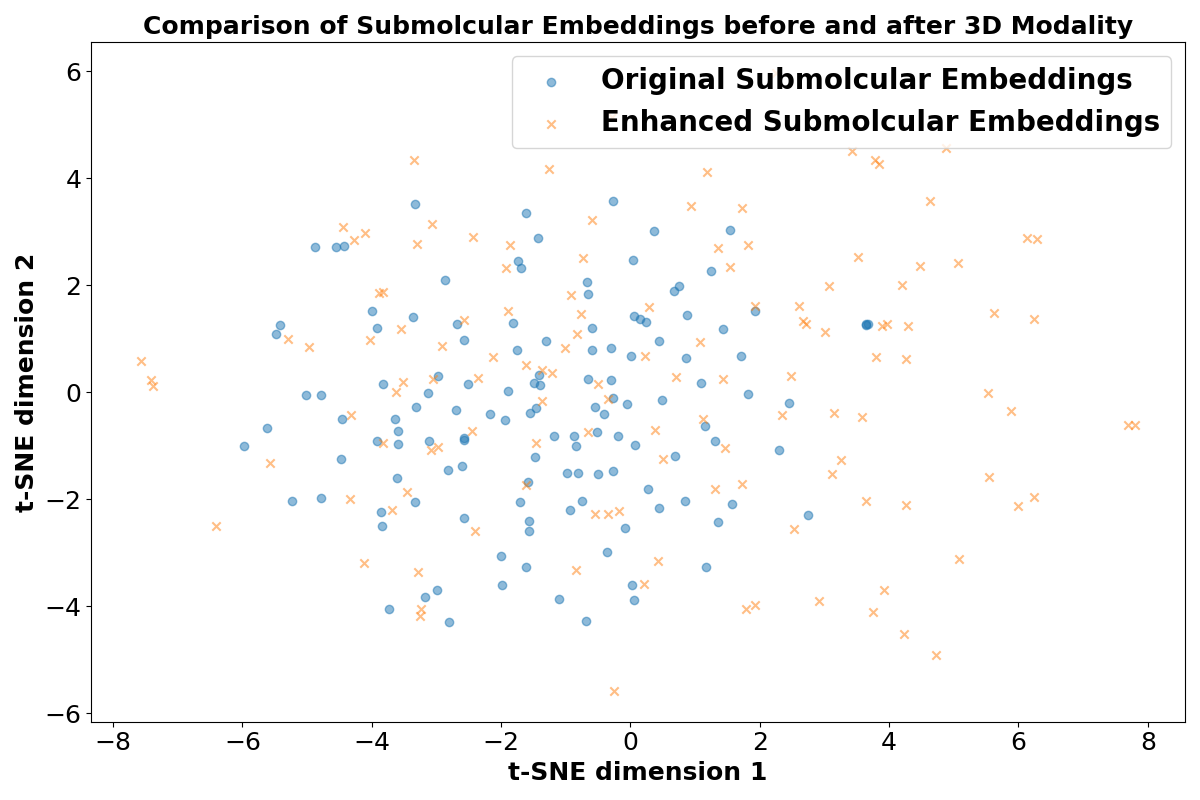}
    \caption{Low dimensional data distribution represented by molecular and substructure embeddings}
    \label{fig:case_2}
\end{figure*}

\begin{figure*}
    \centering
    \includegraphics[width=\textwidth]{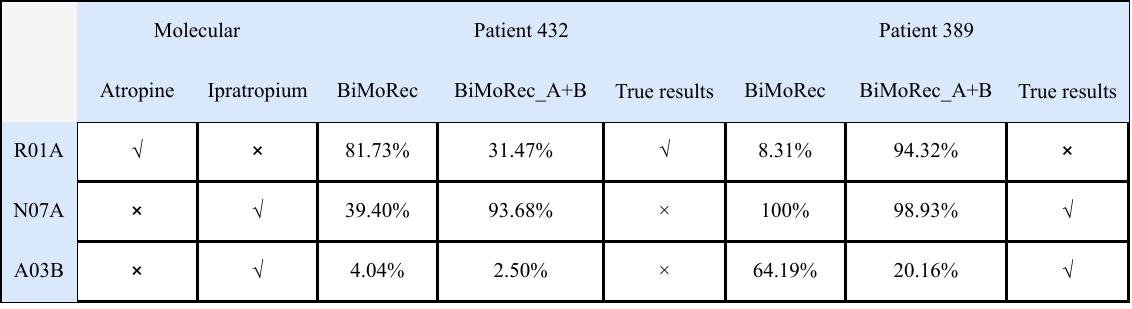}
    \caption{Detailed case studies on molecular 3D modes and multi enhancement in MIMIC-III}
    \label{fig:case_1}
    \vspace{-0.3in}
\end{figure*}


To further clarify the reasoning behind our innovations, we conduct a case study using a sample from MIMIC-III. This specific instance highlights the presence of structural confusion and how it interferes with molecular representation and medication recommendation. It also emphasizes the importance of the interaction between molecular knowledge and individual patient visit representations in identifying key substructures for each visit.

For example, the molecular structures of Atropine and Ipratropium are similar, yet exhibit distinct functional differences. Atropine is primarily used to treat conditions such as blurred vision and arrhythmia, while Ipratropium is primarily used to treat COPD (chronic obstructive pulmonary disease) and other respiratory diseases, administered via nebulization. As shown in Figure~\ref{fig:case_1}, by incorporating the 3D modality, we successfully distinguish between medications such as R01A and A03B, N07A, and recommend the correct medication to the user. As shown in Figure~\ref{fig:case_2}, we analyze the low-dimensional representation of molecular embeddings generated by both the original molecular encoder and the newly introduced bimodal encoder. The more uniform distribution in the PCA (Principal Component Analysis) \citep{pca} plot indicates that significant variability is captured, facilitating molecular differentiation. Similarly, the cluster distribution in the t-SNE (t-distributed Stochastic Neighbor Embedding) \citep{tsne} plot further supports this observation.

In summary, the introduction of 3D modality data addresses structural confusion and effectively optimizes the entire molecular representation space. The subsequent multi-step molecular enhancement mechanism ensures that the optimized representations engage with patient representations at a finer granularity, thereby selecting the medication molecules that the patient truly needs. Case studies validate the above analysis and confirm the effectiveness of our approach.


\section{Conclusion}
This paper presents the BiMoRec medication recommendation model, which addresses the challenge of structural ambiguity through the integration of dual molecular modalities. This approach greatly improves molecular differentiation, optimizing the representation space of molecules. Furthermore, by employing a designed molecular multi-step enhancement mechanism, the interactions between molecules and the relationships between patients and molecules are refined at the visit level. In the final stage, the model optimizes the representation based on the patient's most recent visit, identifying truly effective medication molecules, thereby significantly improving the accuracy and safety of the model. We conducted a series of rigorous experiments on publicly available clinical datasets, validating the effectiveness of our approach.

This study introduces a novel molecular modality and recommendation mechanism, highlighting related challenges and which may provide a foundation for future research. Our future goal is to continue advancing medication recommendation research by exploring large models and their interactions between multiple modalities to optimize our framework, thereby addressing more complex problems and real-world scenarios.

\section*{Declaration of Competing Interest}
The authors declare that they have no known competing financial interests or personal relationships that could have appeared to influence the work reported in this paper.

\printcredits
\section*{Acknowledgments}
\subsection*{Funding}
This work was supported by the Innovation Capability Improvement Plan Project of Hebei Province (No. 22567637H), the S\&T Program of Hebei(No. 236Z0302G), and HeBei Natural Science Foundation under Grant (No.G2021203010 \& No.F2021203038).

\bibliographystyle{cas-model2-names}

\bibliography{BiMoRec}

\bio{}
\endbio

\bio{}
\endbio

\end{document}